\documentclass{article} 
\usepackage[preprint]{colm2026_conference}

\usepackage{booktabs}
\usepackage{array}
\usepackage{caption}
\usepackage{graphicx}
\usepackage{hyperref}
\usepackage{microtype}
\usepackage{multirow}
\usepackage{url}
\usepackage[skins,breakable]{tcolorbox}
\usepackage[dvipsnames]{xcolor}
\usepackage{xspace}
\usepackage{subcaption}
\usepackage{wrapfig}
\usepackage{pifont}
\usepackage{lineno}
\usepackage{enumitem}

\definecolor{Dodgerblue}{RGB}{30,144,255}

\definecolor{f1dark}{RGB}{93,64,55}
\definecolor{f1bg}{RGB}{253,247,245}
\definecolor{f2dark}{RGB}{46,64,87}
\definecolor{f2bg}{RGB}{239,243,247}
\definecolor{f3dark}{RGB}{84,110,122}
\definecolor{f3bg}{RGB}{245,248,250}

\newtcolorbox{caseF}[2]{%
  breakable, enhanced,
  colback=f1bg, colframe=f1dark, boxrule=0.5pt,
  colbacktitle=f1dark, coltitle=white,
  fonttitle=\footnotesize,
  title={\textbf{#1}\enspace\textbar\enspace \textbf{#2}},
  left=6pt, right=6pt, top=4pt, bottom=4pt,
  toptitle=3pt, bottomtitle=3pt,
}

\newcommand{\ours}{\textsc{CocoaBench}\xspace}
\newcommand{\scaffold}{\textsc{Cocoa-Agent}\xspace}
\newcommand{\cmark}{\ding{51}}
\newcommand{\xmark}{\ding{55}}

\title{\ours: Evaluating unified digital agents in the wild}

\author{\ours team}

\begin{document}
\maketitle

\begin{abstract}

LLM agents now perform strongly in software engineering, deep research, GUI automation, and various other applications, while recent agent scaffolds and models are increasingly integrating these capabilities into unified systems. Yet, most evaluations still test these capabilities in isolation, which leaves a gap for more diverse use cases that require agents to combine different capabilities. We introduce \ours, a benchmark for unified digital agents built from human-designed, long-horizon tasks that require flexible composition of \textbf{vision}, \textbf{search}, and \textbf{coding}. Tasks are specified only by an instruction and an automatic evaluation function over the final output, enabling reliable and scalable evaluation across diverse agent infrastructures. We also present \scaffold, a lightweight shared scaffold for controlled comparison across model backbones. Experiments show that current agents remain far from reliable on \ours, with the best evaluated system achieving only 45.1\% success rate. Our analysis further points to substantial room for improvement in reasoning and planning, tool use and execution, and visual grounding.\footnote{Project page: \url{https://cocoabench.github.io/}}


\end{abstract}

\section{Introduction}
LLM agents are showing strong potential across an expanding set of domains, including software engineering~\citep{yang2024sweagent}, GUI automation~\citep{wang2025opencua}, and report generation with deep research~\citep{openai2025deepresearch}. 
Recent agentic frameworks, e.g., OpenClaw~\citep{openclaw2026repo} and Claude Cowork~\citep{anthropic2026cowork}, as well as models, e.g., GPT-5.4~\citep{openai2026gpt54}, Claude Sonnet 4.6~\citep{anthropic2026claudesonnet46}, and Seed-2.0~\citep{bytedanceseed2026seed2model}, aim to unify these capabilities into a single system, moving toward a unified digital agent that can assist humans with complex tasks. 
However, existing benchmarks still largely focus on a single domain or a single interaction mode (e.g., CLI-only~\citep{jimenez2024swebench}, GUI-only~\citep{xie2024osworld}, or predefined tool APIs~\citep{li2026the}), making them insufficient for systematically evaluating unified agent capabilities on diverse tasks in open environments.

\begin{figure}[t]
    \centering
    \includegraphics[width=\textwidth]{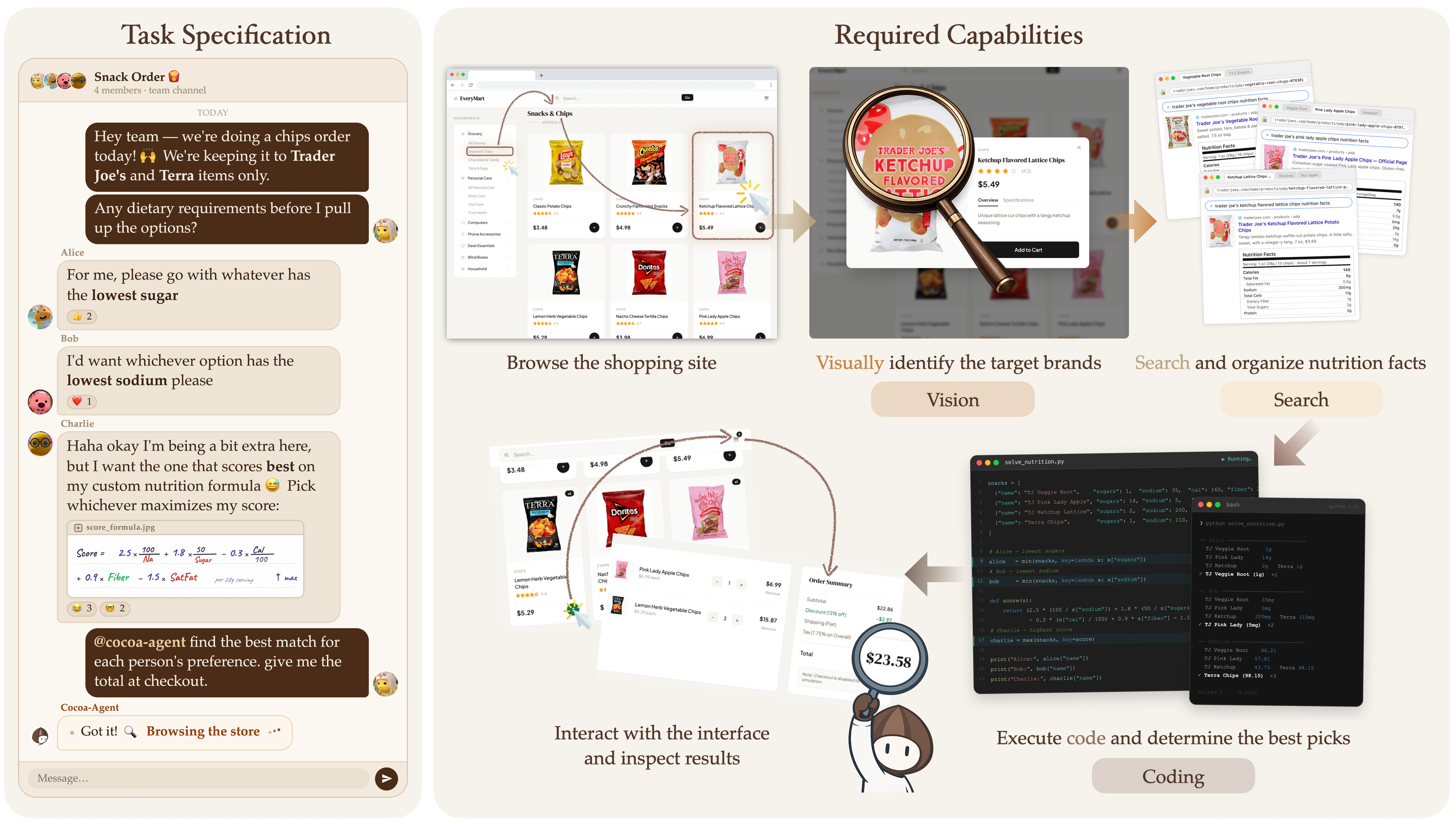}
    \caption{\ours evaluates agents on complex digital tasks that require flexible composition of core capabilities such as vision, search, coding. The shopping example shown here illustrates one such task and highlights the multi step, compositional nature of the benchmark.}
    \label{fig:intro-figure1}
\end{figure}

\ours is designed to evaluate general purpose digital agents on complex tasks that require composing multiple core capabilities. We focus on three fundamental capabilities that are essential for a strong digital agent: 
(1) \textbf{coding} (or, more broadly, terminal use), which enables code-based problem solving, supports quantitative analysis, and allows agents to invoke structured tools and APIs; 
(2) \textbf{search}, which enables information seeking, navigation, and synthesis across online sources; and 
(3) \textbf{vision}, which enables agents to interpret visual inputs and interact with GUIs. 
Beyond mastering each capability in isolation, a strong agent must also \textit{plan effectively and compose these capabilities adaptively} to achieve a target goal (Figure~\ref{fig:intro-figure1}). 

\ours tasks are specified minimally by an instruction and an evaluation function over the agent's final output, without being tied to a particular runtime, interface, or tool ecosystem. This design keeps the benchmark agnostic to specific agent infrastructures and requires agents to reason about tool use in an open world setting, rather than operate within pre-specified apps. To make evaluation reliable without sacrificing task complexity, we equip each task with an evaluation script, without relying on LLM judges or human evaluation. For action centric tasks, where correctness depends on multi-step interaction, we design outcome based proxy evaluators whose success strongly implies correct execution. This process to outcome transformation preserves open ended workflows while keeping evaluation reproducible and scalable.

We evaluate \ours in two settings: (1) using existing agent products as complete systems, and (2) under \scaffold, a lightweight shared scaffold that enables more controlled comparison across backbone models. Our experiment shows that the best-performing agent (GPT-5.4 under Codex) achieves a success rate of only 45.1\%, and leading open-source models such as Kimi-k2.5 and Qwen3.5 reach only 11.8\% and 9.8\% respectively, highlighting significant room for improvement in current agent capabilities.
We also find that scaffold design plays an important role. Coding-oriented scaffolds such as Codex and Claude Code generalize well beyond their original domain, serving as effective task solvers on \ours.
Analysis of tool usage reveals that top-performing models allocate more of their actions to code execution, indicating that programmatic processing is an effective strategy for the multi-step reasoning and structured output formatting that \ours tasks demand.
Our error analysis shows that current systems remain unreliable along three key dimensions of unified digital agency: reasoning and planning, tool interaction and execution, and visual grounding.
We open-source \ours, including all task instructions, evaluation scripts, and the full implementation of \scaffold scaffold, to facilitate reproducible evaluation and future research on general purpose digital agents.

\section{Related Work}
\subsection{Evaluating digital agents}

As LLM-based agents expand from single-domain tools to general-purpose digital assistants, the need for comprehensive evaluation benchmarks has grown accordingly.
Table~\ref{tab:benchmark-comparison} summarizes representative agent benchmarks that are widely used in recent frontier-model (e.g., Gemini-3.1 pro, GPT-5.4 and Claude-Opus-4.6) evaluations till March 2026.
We compare these benchmarks along application focus, infrastructure coupling, reward verifiability, and required core capabilities (vision, search, and coding).

Existing agent benchmarks each capture a useful but limited slice of digital agent evaluation. 
OSWorld~\citep{xie2024osworld} studies real computer use in VM-based desktop environments with task-specific setup; prior analysis suggests that GUI grounding and operational knowledge are major bottlenecks, while complex reasoning demands are relatively limited. 
SWE-bench Pro~\citep{deng2025swe} and TerminalBench-2 focus on repository issue resolution and CLI task execution, respectively, but both are largely restricted to software-engineering domains. 
MCP Atlas~\citep{bandi2026mcp} and Tool Decathlon~\citep{li2026the} broaden coverage to tool-use settings, yet they still operate within fixed tool ecosystems, emphasizing tool understanding and execution over open-ended strategy. 
BrowseComp~\citep{wei2025browsecomp} targets open-web research, but follows a fairly specific pattern of iterative search, candidate generation, and answer verification. 
GDPval~\citep{patwardhan2025gdpval} covers professional work across 44 occupations, but this realism makes evaluation harder: its main metric is blinded expert pairwise judgment of the deliverables (70.8\% human inter-rater agreement). 
Unlike these benchmarks, \ours targets a different balance: it does not assume a fixed runtime or tool ecosystem. 
Instead, each task is specified by an instruction and an evaluation function over final outputs, while task design explicitly requires composing vision, search, and coding across diverse digital tasks.

\begin{table}[t]
    \centering
    \small
    \setlength{\tabcolsep}{2.8pt}
    \begin{tabular}{p{2.8cm}p{3.8cm}c c >{\centering\arraybackslash}p{0.42cm} >{\centering\arraybackslash}p{0.42cm} >{\centering\arraybackslash}p{0.42cm}@{}}
        \toprule
        \multirow{2}{*}{\textbf{Benchmark}} & \multirow{2}{*}{\textbf{Application focus}} & \multirow{2}{*}{\textbf{Infra coupling}} & \multirow{2}{*}{\textbf{\shortstack{Verif.\\reward}}} & \multicolumn{3}{c}{\textbf{Capabilities}} \\
         &  &  &  & \textbf{V} & \textbf{S} & \textbf{C} \\
        \midrule
        OSWorld & Computer use & VM + task setup & \cmark & \cmark & \xmark & \xmark \\
        BrowseComp & Web research & Open & \cmark & \xmark & \cmark & \xmark \\
        SWE-bench Pro & Repository-issue resolution & Repo container & \cmark & \xmark & \xmark & \cmark \\
        TerminalBench-2 & CLI task execution & Task container & \cmark & \xmark & \xmark & \cmark \\
        MCP Atlas & MCP server orchestration & Fixed apps & \cmark & \xmark & \xmark & \cmark \\
        ToolDecathlon & Cross-app tool using & Fixed apps & \cmark & \xmark & \xmark & \cmark \\
        GDPval & Occupation-grounded tasks & Open & \xmark & \cmark & \cmark & \cmark \\
        \ours & Diverse digital tasks & Open & \cmark & \cmark & \cmark & \cmark \\
        \bottomrule
    \end{tabular}

    \caption{Comparison of representative agent benchmarks by application focus, infrastructure coupling, reward verifiability, and core capability coverage (V=Vision, S=Search, C=Coding).}
    \label{tab:benchmark-comparison}
\end{table}

\subsection{General Digital Agents}
Large language model based agents have become capable of performing complex tasks across different digital environments, but existing systems typically operate within a single interaction modality.
SWE-Agent~\citep{yang2024sweagent} and OpenHands~\citep{wang2025openhandsopenplatformai} target software engineering, while Codex, Claude Code, and Terminus-2~\citep{merrill2026terminalbenchbenchmarkingagentshard} operate in terminal environments. On the visual side, Aguvis~\citep{xu2025aguvisunifiedpurevision}, OpenCUA~\citep{wang2025opencua}, and UI-TARS~\citep{wang2025uitars2technicalreportadvancing} enable agents to operate graphical interfaces through screenshot understanding and coordinate-based actions. Deep research agents~\citep{openai2025deepresearch} address yet another axis, performing multi-step web search and synthesis.
While effective within their respective domains, these systems each rely on a single interaction modality and do not generalize across capability boundaries.

Recent systems such as OpenClaw~\citep{openclaw2026repo} and ChatGPT Agent~\citep{openai2025introducingchatgptagent} aim to integrate browsing, coding, and visual interaction into a single agent, but systematic evaluation of such general-purpose agents remains challenging, as existing benchmarks typically assess only a subset of the required capabilities. \ours and \scaffold are designed to address this gap, providing tasks that explicitly require the composition of vision, search, and coding alongside a lightweight agent framework with integrated sandbox support for reproducible evaluation.

\section{CocoaBench}
\subsection{Task construction}


\ours consists of 153 human-authored tasks designed to evaluate unified agents on complex problem solving. 
We first identified practical scenarios in which agents are expected to provide assistance, covering research, entertainment, shopping, business, and other everyday tasks. 
For each scenario, we instantiated 3 to 5 concrete tasks to form the final \ours dataset. 
Task authors adhered to three key criteria:
\begin{itemize}[leftmargin=1.2em, labelsep=0.4em]
    \item Each task should require the integration of multiple capabilities.
    \item Each task should pose a nontrivial challenge for humans in realistic settings.
    \item Dependencies on external resources should remain stable over time, so that task validity is not compromised by changes in third-party content.
\end{itemize}

\paragraph{Inclusive task settings.}
Unlike benchmarks that are tightly coupled to specific environments like OSWorld~\citep{xie2024osworld} or fixed tool ecosystems like Tool Decathlon~\citep{li2026the}, each task in \ours is \textbf{minimally} defined and specified by an instruction and an evaluation function over agent outputs. 
For tasks requiring multimodal inputs or additional resources, we host the required assets online and include their URLs directly in the task instructions.
This design makes the benchmark compatible with diverse agent infrastructures, including locally deployed ones, e.g., OpenClaw~\citep{openclaw2026repo}, and the hosted or sandboxed ones, e.g., ChatGPT Agent Mode~\citep{openai2026gpt54}. 
It also allows tasks to be more diverse, rather than being constrained by specific environments or infrastructures. 
Furthermore, it evaluates whether agents can reason about which tools to use for each task in an open-world setting, instead of selecting only from a fixed, predefined toolset. 

\paragraph{Automatic evaluation functions.} 
Each task is paired with its own evaluation function, enabling automatic and reproducible assessment. 
Whenever possible, we require outputs in a unique structured format, e.g., \texttt{str}, \texttt{list}, or \texttt{dict}. 
In some real-world tasks, agents must take actions with an environment to complete the task, beyond answering questions. Directly verifying the action sequence, however, is often impractical. 
We therefore use proxy outcome verifiers based on automatically checkable end results, designed so that a correct outcome is unlikely without successful execution. 
For example, in the shopping task shown in Figure~\ref{fig:intro-figure1}, we verify the final price returned by the agent, since obtaining the correct value typically requires both correct website interaction and correct reasoning over the user request. This enables scalable evaluation while retaining realistic and diverse task settings.


\paragraph{Quality control.} To ensure quality, all tasks, reference answers, and evaluation functions underwent a rigorous peer-review process before inclusion. Reviewers verified that instructions were unambiguous, output formats were well-defined, and reference answers were correct. Furthermore, they ensured that tasks did not allow for trivial shortcuts that bypass the intended reasoning or interaction process. We also confirmed that external resources remain accessible to support reproducibility. Additionally, we conducted pilot experiments with several agents on an initial version of the benchmark. By inspecting agent logs, we identified recurring failure patterns and distinguished agent failures from design issues. Tasks with persistent ambiguity were removed, and this iterative refinement process substantially enhanced the final quality of the dataset.

\subsection{Task diversity and composition}

\begin{figure}[t]
    \centering
    \includegraphics[width=\textwidth]{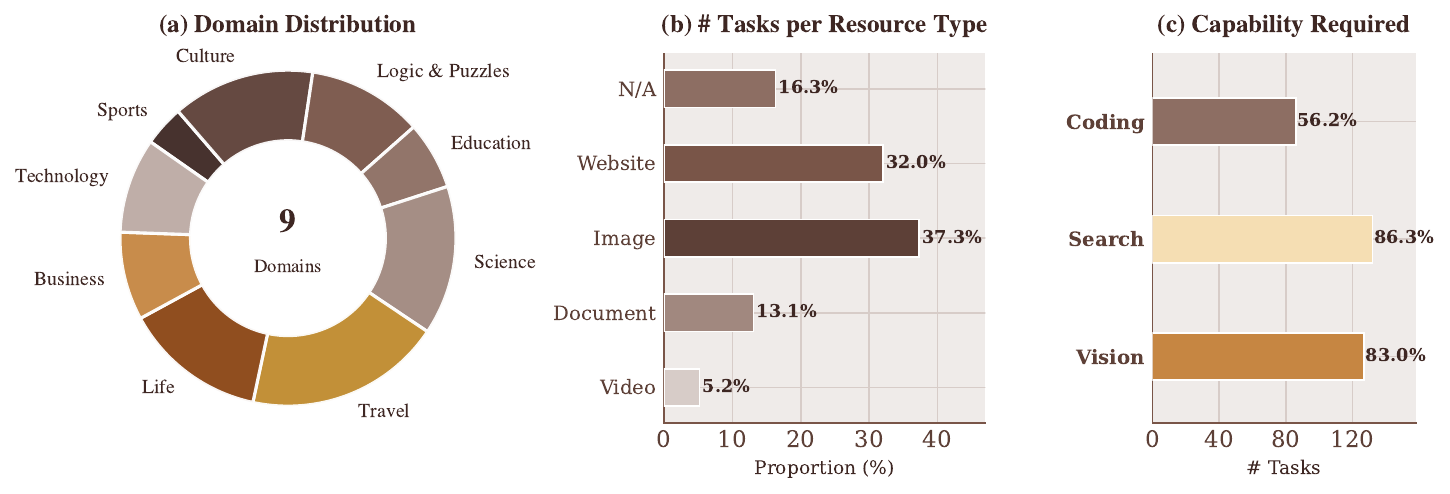}
    \caption{\textbf{Statistics of \ours.}
    (\textbf{a}) Distribution of task domains, covering a wide range of everyday topics.
    (\textbf{b}) Distribution of resource types required by the tasks. Documents include \texttt{.csv}, \texttt{.pdf}, and \texttt{.bibtex}.
    (\textbf{c}) Human-annotated key capabilities required for each task, including Vision, Search, and Coding. The Vision, Search, and Coding types are not mutually exclusive; numbers on the bars indicate the proportion of all tasks, and the x-axis represents the number of tasks.}
    \label{fig:composition}
\end{figure}

\paragraph{Task domains.} 
\ours consists of tasks spanning 9 diverse domains (Figure~\ref{fig:composition}(a)), including Business, Culture, Education, Life, Logic \& Puzzles, Science, Sports, Technology, and Travel.
These scenarios closely mirror everyday challenges that can often be solved, or even optimally solved, given sufficient time and patience, through careful planning and deliberate navigation of task resources and the broader digital world.

\paragraph{Task resources.}
The tasks in \ours are supported by diverse resources, which are either carefully collected from the internet or provided by the task designers. 
These include webpages, videos, images, and documents in realistic environments. Notably, task designers hosted 17 websites and contributed artifacts such as Weights \& Biases logs, their own ChatGPT conversations, and even a collection of Costco receipts, all of which help construct realistic and challenging tasks.
We have made our best effort to ensure that these resources are easily acquired by anyone and remain stable. 
The distribution of resource types is shown in Figure~\ref{fig:composition} (b).

\begin{wrapfigure}{r}{0.45\textwidth}
    \centering
    \vspace{-8em}
    \includegraphics[width=0.43\textwidth]{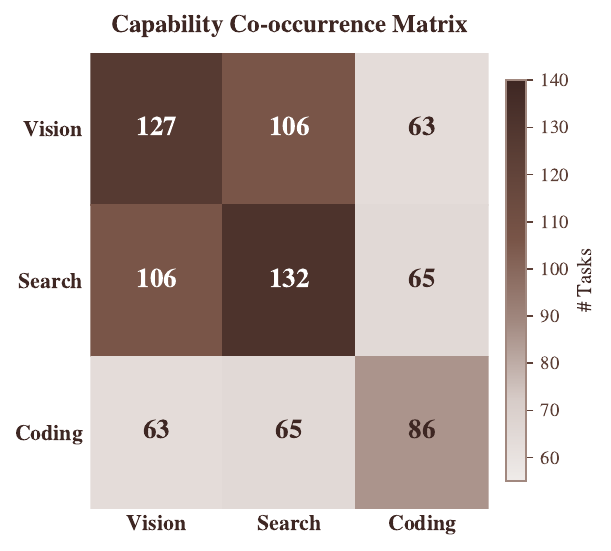}
    \vspace{-0.5em}
    \caption{Co-occurrence matrix of required capabilities (Vision, Search, and Coding).}
    \label{fig:ability-cooccurrence}
    \vspace{-2em}
\end{wrapfigure}

\paragraph{Target capabilities.}
We categorize the key capabilities required to solve each task into three main types: \textbf{Vision}, \textbf{Search}, and \textbf{Coding}. 
The primary labels are determined based on human annotation. 
As shown in Figure~\ref{fig:composition} (c), a task is labeled as \textit{Vision} if visual information must be extracted to solve it correctly; 
\textit{Search} if accessing and analyzing information from the internet is necessary; and 
\textit{Coding} if writing code is considered important for solving the task efficiently and reliably. 
Notably, 98\% of tasks require multiple capabilities, and their co-occurrence matrix is illustrated in Figure~\ref{fig:ability-cooccurrence}. Interestingly, although Coding is annotated as important for 56.2\% of tasks, our later analysis in \hyperref[sec:tool-statistics]{Section~\ref*{sec:tool-statistics}} shows that stronger agents rely on code execution even more broadly than expectation.

\section{Experiment settings}

\subsection{Existing agentic systems}

We evaluate representative agent systems on \ours to cover a range of agent designs and capability profiles. 
(1) \textbf{ChatGPT Agent Mode}~\citep{openai2025introducingchatgptagent} is one of the earliest unified digital agents, with support for browsing, coding, and visual interaction in a sandbox environment. 
(2) \textbf{OpenClaw}~\citep{openclaw2026repo} is an open source framework for unified digital agency that can be deployed on personal computers. 
We instantiate it with GPT-5.4 thinking high~\citep{openai2026gpt54} and Claude Sonnet 4.6 thinking high~\citep{anthropic2026claudesonnet46} as backbones. 
(3) \textbf{Codex}\footnote{\url{https://openai.com/codex/}} and (4) \textbf{Claude Code}\footnote{\url{https://code.claude.com/docs/en/overview}} are two representative coding agent products. 
We also use GPT-5.4 thinking high and Claude Sonnet 4.6 thinking high as the backbone model, respectively.
(5) \textbf{OpenAI Deep Research}~\citep{openai2025deepresearch} is included as a research-oriented agent for long horizon web information seeking and synthesis. 
We use the \texttt{o4-mini} version of it. 
Unless otherwise specified, each run uses a 30 minute wall clock budget with a maximum of 50 interaction turns.

\subsection{\scaffold}

Similar to the Bash-Only setting in SWE-Bench~\citep{jimenez2024swebench}, although our full leaderboard compares arbitrary agentic systems, we also aim to compare the agentic capabilities of LLMs under a shared agent framework.
We develop \scaffold, a unified agent scaffold that is intentionally lightweight and modular, allowing us to control agentic components and make comparisons more analytically interpretable.
It is built on top of AIO Sandbox\footnote{\url{https://github.com/agent-infra/sandbox}}, an all-in-one sandbox runtime that integrates browser, shell, file system within a single Docker container. 
\scaffold adopts a ReAct-based scaffold, equipping model backbones with general purpose tools for browser interaction (both DOM-level APIs and screenshot-based GUI control), terminal execution, file manipulation, and code execution. 
With the integration of sandbox, it also enables safer execution and more scalable parallel evaluation. 
We also expect it to provide a practical foundation for future research on reinforcement learning for unified digital agents. 

To evaluate model backbones under a consistent agent scaffold, we include: (1) \textbf{Claude Sonnet 4.6 (thinking high)}, (2) \textbf{GPT-5.4 (thinking high)}, (3) \textbf{Gemini-3.1-pro (thinking high)}, and (4) \textbf{Gemini-Flash-3.0}. We additionally include strong open-source multimodal models: (5) \textbf{Kimi-k2.5}~\citep{moonshotai2026kimmik25}, an MoE model with 1T parameters and 32B active parameters, and (6) \textbf{Qwen3.5-397B-A13B}~\citep{qwenteam2026qwen35}, an MoE model with 397B parameters and 13B active parameters. Together, these choices cover both leading proprietary models and strong open-source multimodal alternatives.

\section{Results and analysis}

\subsection{Overall results}

\begin{figure}[t]
    \centering
    \vspace{-2em}
    \includegraphics[width=0.9\linewidth]{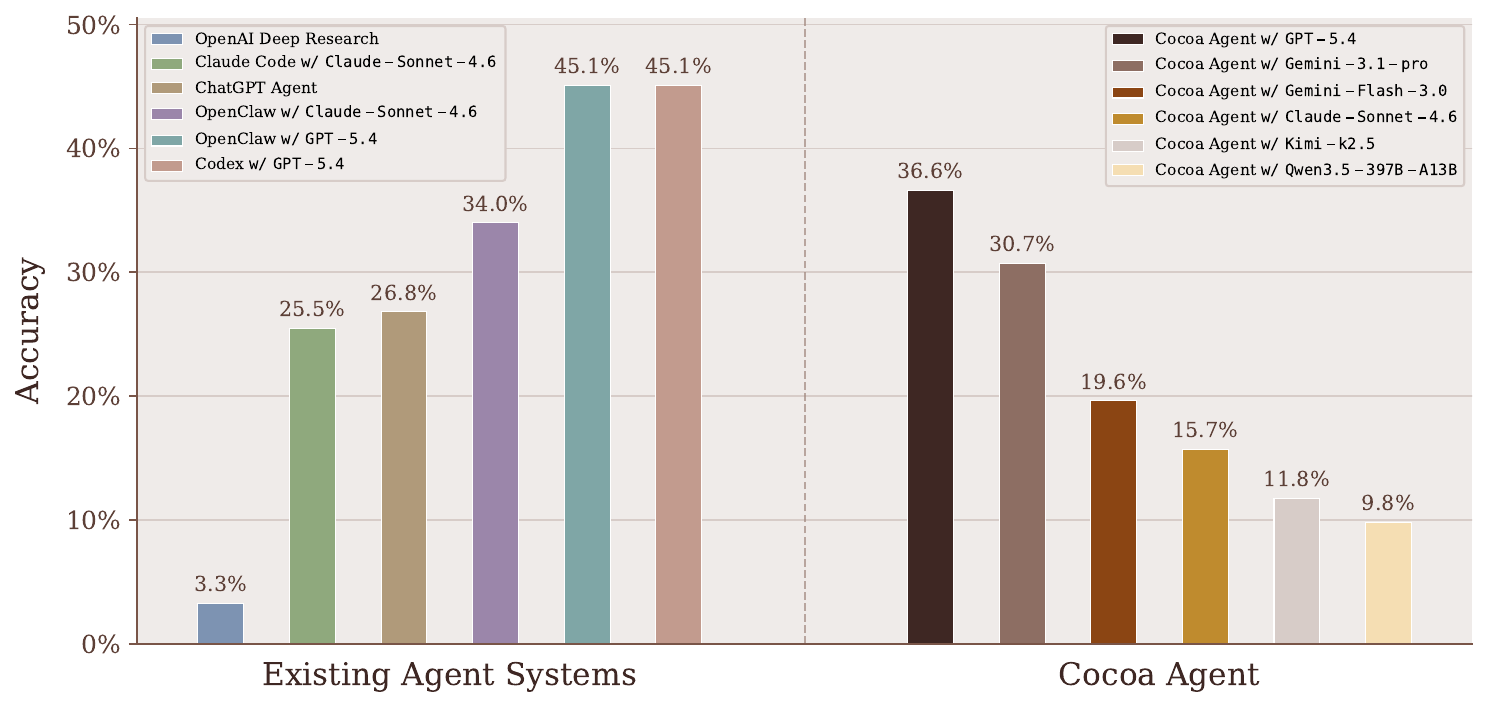}
    \caption{Overall performance on \ours for representative agent systems and model backbones under the shared \scaffold scaffold.
    }
    \label{fig:overall_results}
\end{figure}

We first report the main results (accuracy) on \ours for both representative existing agent systems and model backbones instantiated under \scaffold. 
Figure~\ref{fig:overall_results} shows two complementary views: the left compares complete agent systems in different scaffolds, while the right compares diverse backbones under a shared \scaffold scaffold.

From the model perspective, GPT 5.4 is the most consistently strong backbone across scaffolds. 
It attains 45.1\% under both Codex and OpenClaw, and still reaches 36.6\% under \scaffold, corresponding to the top three entries in the leaderboard. 
Claude Sonnet 4.6 can also be competitive, achieving 34.0\% under OpenClaw, but its performance is less stable across other scaffolds, dropping to 25.5\% in Claude Code and 15.7\% in \scaffold. 
By contrast, the open source models remain clearly behind the leading proprietary models, with Kimi k2.5 and Qwen3.5 397B A13B reaching 11.8\% and 9.8\%, respectively. 
Overall, these results suggest that backbone quality still matters substantially, with GPT 5.4 standing out as the most robust model on \ours.

The agent scaffold also plays a crucial role. 
A notable finding is that scaffolds originally developed for coding, including CodeX and Claude Code, can already act as fairly general problem solvers on \ours. 
OpenClaw also appears to be a robust scaffold, yielding strong results with both GPT 5.4 and Claude Sonnet 4.6. 
While \scaffold is not the strongest-performing scaffold, it already attains reasonably strong performance with capable backbones, making base model comparisons meaningful. 
Given its simplicity and integrated sandbox, we believe it also serves as a promising baseline for future research on unified digital agents, including data engineering and reinforcement learning-based training.

\subsection{Cost and model performance}

\begin{figure}[t]
    \centering
    \includegraphics[width=\linewidth]{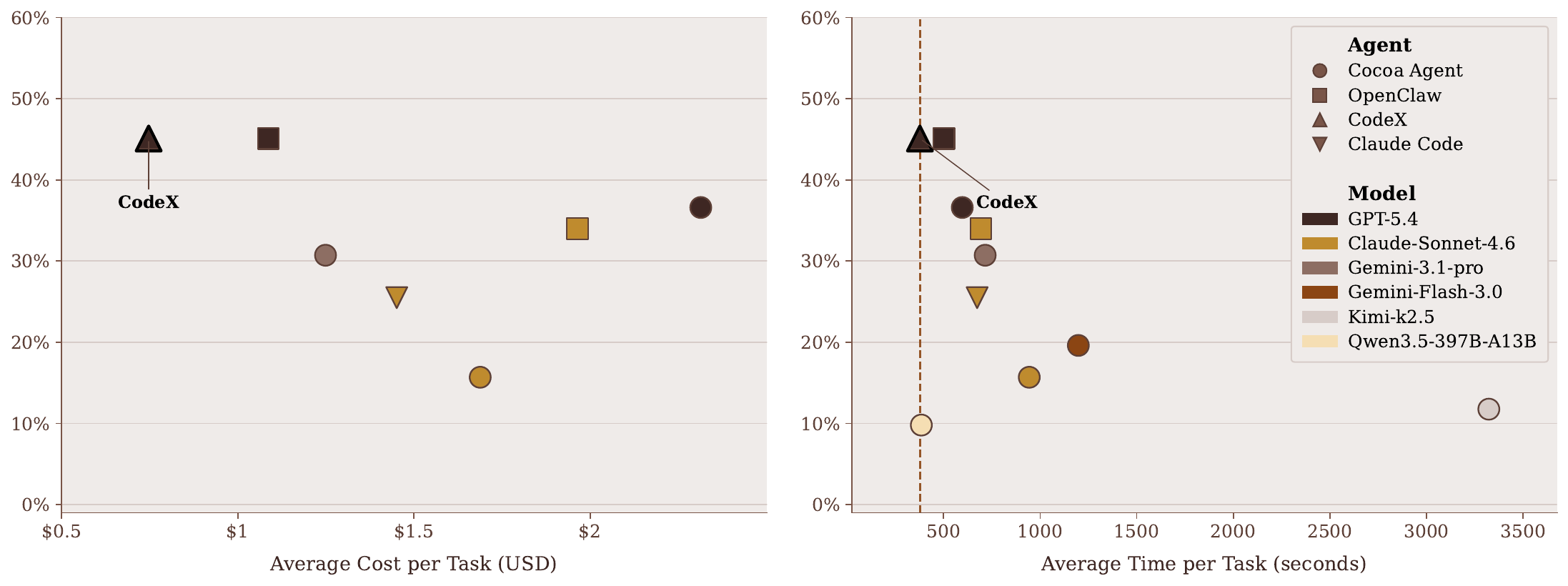}
    \caption{Accuracy-cost (left) and accuracy–time (right) trade-offs across agents. Marker shapes denote agent/scaffold types and colors denote base models.}
    \label{fig:cost_comparison}
\end{figure}

We compare model performance against average cost and task completion time. The average cost per task ranges from \$0.5 to \$2.5, while average completion time ranges from 380s to 3400s. As shown in Figure~\ref{fig:cost_comparison}, both comparisons show a consistent trend: CodeX achieves the best balance between cost efficiency and performance and lies on the Pareto frontier, while the other agents show no clear complementary advantages. This suggests that higher monetary or time costs do not necessarily lead to better performance. For example, \scaffold w/ Qwen3.5-397B-A13B has a completion time comparable to that of CodeX but achieves 35.3\% lower accuracy.
Cost efficiency also depends strongly on the scaffolding design. Even with GPT-5.4 as the base model, Codex costs \$0.75 per task, compared with \$1.09 for OpenClaw and \$2.31 for \scaffold.

\subsection{Tool statistics}\label{sec:tool-statistics}

We further analyze the tool calls recorded under \scaffold across the six evaluated models to examine how different backbones utilize the available tools and compose core capabilities during task solving.

\begin{wrapfigure}{r}{0.5\textwidth}
    \centering
    \vspace{-1em}
    \includegraphics[width=\linewidth]{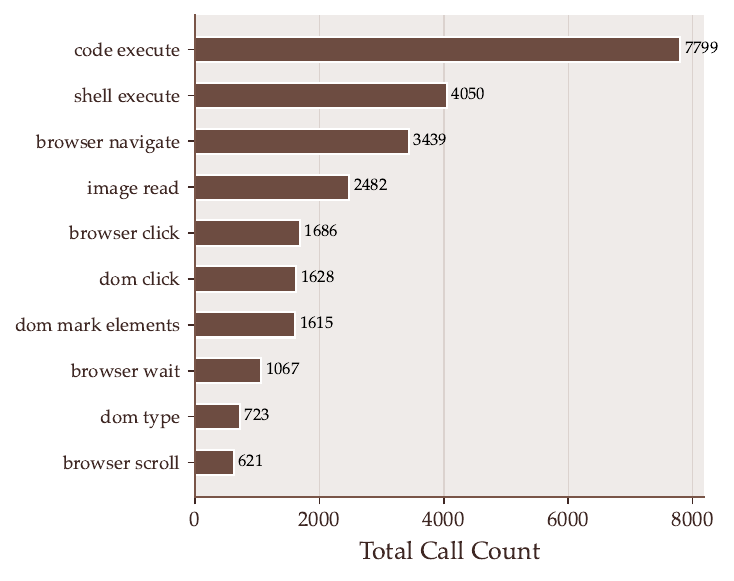}
    \caption{Top 10 most frequently used tools, ranked by total call count.}
    \label{fig:tool_ranking}
\end{wrapfigure}

\begin{figure}[t]
    \vspace{-1em}
    \centering
    \includegraphics[width=0.8\linewidth]{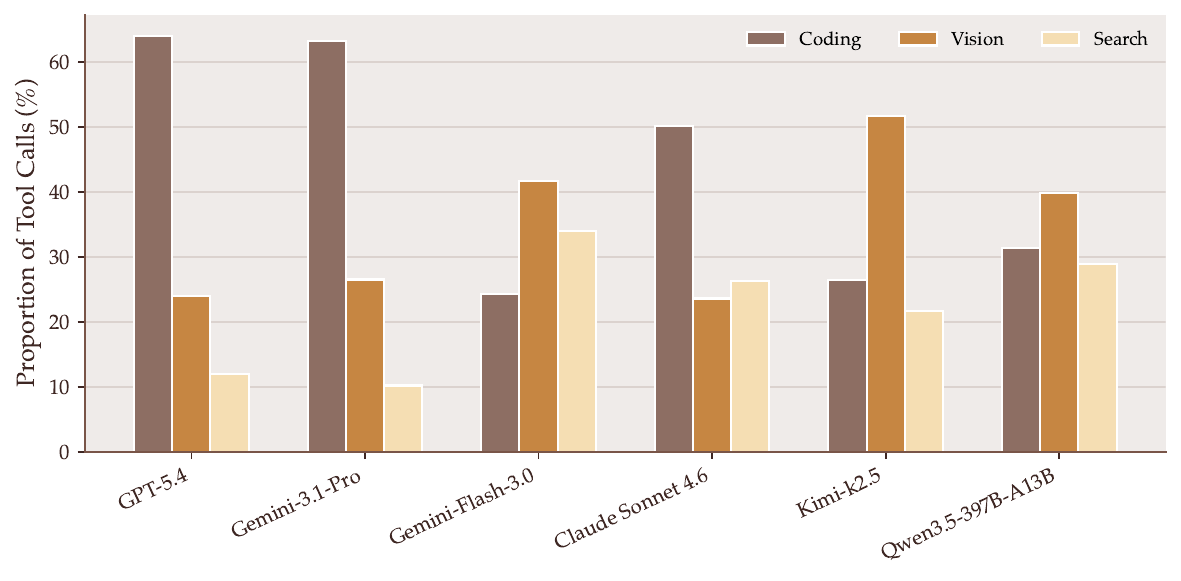}
    \vspace{-1em}
    \caption{Per-model distribution of tool calls across the three capability categories under \scaffold.}
    \label{fig:category_distribution}
    \vspace{-1em}
\end{figure}




\paragraph{Tool call distribution.}
Figure~\ref{fig:tool_ranking} reports the aggregate call counts of the ten most frequently used tools across the 6 models under \scaffold. 
Coding tools dominate overall: \texttt{code\_execute} and \texttt{shell\_execute} together account for the largest share of total invocations, followed by browser level actions such as \texttt{browser\_navigate} and \texttt{image\_read}, with DOM-level interaction tools appearing at moderate frequency. 
This distribution reflects the compositional demands of \ours tasks, which generally require agents to both perceive and acquire information from diverse online sources and to process and synthesize that information into structured outputs. 
The prominence of coding tools suggests that programmatic execution is central to solving \ours tasks, providing a reliable approach for multi-step reasoning, data processing, and structured output formatting.


\vspace{-8pt}

\paragraph{Tool usage and performance.}
To compare tool usage profiles across models, we map each tool to one of the three key capabilities defined in \ours: \textit{vision}, \textit{search}, \textit{coding}, as detailed in Table~\ref{tab:cocoa-agent-tools}. 
As shown in Figure~\ref{fig:category_distribution}, models differ substantially in their tool usage. 
GPT-5.4 and Gemini~3.1~Pro allocate over 60\% of their tool calls to coding tools, using browser interaction primarily for information acquisition. 
Kimi-k2.5 and Gemini-Flash-3.0 exhibit the opposite profile: Kimi-k2.5 assigns 51.7\% of its calls to vision tools, while Gemini-Flash-3.0 directs 34.0\% toward DOM-level search operations. This divergence in tool usage is reflected in task performance. 
GPT-5.4 (64.0\% coding, 36.6\% SR) and Gemini~3.1~Pro (63.2\% coding, 26.1\% SR) achieve the highest success rates, whereas Kimi-k2.5 (26.4\% coding, 11.8\% SR) and Qwen3.5-397B-A13B (31.3\% coding, 9.8\% SR) allocate near 30\% of their tool calls to coding and rank at the bottom. 
This pattern suggests that code execution serves a dual role in \ours: as an efficient action space that reduces the number of interaction steps required per subtask, and as an analytical tool that enables complex reasoning over gathered information. 
Stronger models leverage this by separating information acquisition (via vision and search) from downstream processing (via code), whereas weaker models underutilize programmatic processing and remain in the browser for both phases.

\begin{figure}[t]
    \centering
    \vspace{-1.5em}
    \includegraphics[width=0.98\linewidth]{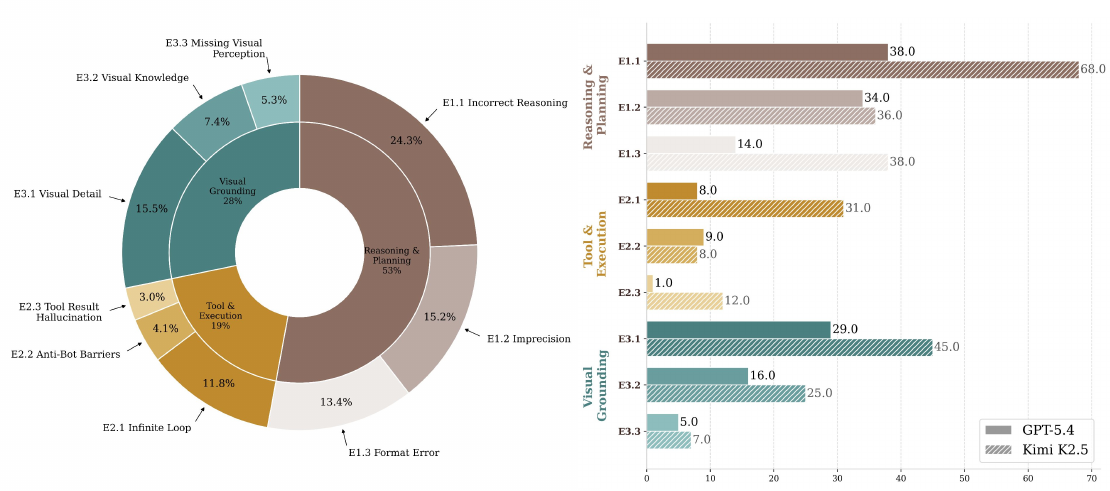}
    \caption{
        \textbf{Left:} Error type distribution across all six models evaluated on \ours using the \scaffold framework (based on 712 failure trajectories).
        \textbf{Right:} Comparisons of error distributions between GPT-5.4 and Kimi K2.5.
    }
    \label{fig:failure_taxonomy_breakdown-cocoa}
\end{figure}

\subsection{Error analysis}
To understand where agents fail on \ours and what these failures reveal about the challenges of building unified digital agents, we conduct a structured error analysis over all six evaluated models, covering 712 failure trajectories out of 918 total task attempts. Failure causes are annotated by an LLM as judge (Claude Sonnet 4.6, grade prompt is provided in Appendix~\ref{app:judge-prompt}). We organize the annotated error types into three classes. \textbf{Reasoning \& Planning} (E1) describes cases where agents fail to devise an effective approach, reason imprecisely about crucial details, or lose track of task requirements such as the requested output format.
\textbf{Tool \& Execution} (E2) focuses on how the agent interacts with tools and interfaces, especially whether it can execute the right steps, in the right order, and recover when execution goes off track.
\textbf{Visual Grounding} (E3) captures cases where agents fail to properly perceive or interpret visual information, such as overlooking subtle but task critical details, confusing interface elements, or misreading visual content. The overall error distribution across these categories is shown in Figure~\ref{fig:failure_taxonomy_breakdown-cocoa} (left). Full category definitions and concrete examples are provided in Appendix~\ref{app:error-taxonomy}.


To better understand the gap between the leading agent model and other models, we compare the error distributions of GPT 5.4 and Kimi K2.5 (Figure~\ref{fig:failure_taxonomy_breakdown-cocoa}, right). Relative to GPT 5.4, Kimi K2.5 fails more often on (E1.1) incorrect reasoning, suggesting weaker procedural knowledge for handling diverse scenarios. It also exhibits a substantially higher rate of format errors (E1.3), indicating that over long interaction horizons, it is more likely to lose track of instructions introduced earlier in the trajectory. In terms of tool use, Kimi K2.5 is more prone to infinite loops (E2.1): when tool outputs are unexpected, it is more likely to get stuck in repetitive tool calls and fail to recover. Finally, it underperforms noticeably on visual grounding errors, especially (E3.1) visual detail, indicating that it is less reliable at noticing fine grained visual information.

\section{Conclusion}
\ours is designed to evaluate unified digital agents beyond isolated capability tests, focusing instead on whether agents can flexibly compose vision, search, and coding to solve complex digital tasks. Across both end to end agent systems and controlled evaluations under the shared \scaffold scaffold, our results show that current systems still struggle to solve \ours reliably. Our analysis further suggests that coding is an important ingredient in strong agent performance, while error analysis reveals crucial weaknesses in planning and reasoning, tool use and execution, and visual grounding, which point to several promising directions for future work. We hope \ours and \scaffold can serve as useful foundations for future research on more capable general purpose digital agents.

\bibliography{colm2026_conference}
\bibliographystyle{colm2026_conference}

\appendix
\clearpage
\section{Contributors}

\noindent\textbf{Data Curation}

Zhining Zhang$^{1}$, Tianyang Liu$^{1}$, Yuheng Zha$^{1}$, Qiyue Gao$^{1}$, Jixuan Chen$^{1}$, Hexi Jin$^{1}$, Boyuan Zheng$^{1}$, Shibo Hao$^{1}$

\noindent\textbf{Additional Task Authors}

Yijiang Li$^{1}$, Tommaso Cerruti$^{5}$, Licheng Liu$^{6}$, Zhifei Li$^{4}$, Zhengtao Han$^{4}$, Pracha Promthaw$^{1}$

\noindent\textbf{Infrastructure}

Zhiqi Liang$^{1}$, Junli Wang$^{1}$, Zilong Wang$^{1}$

\noindent\textbf{Evaluation}

Haoxiang Zhang$^{1}$, Hexi Jin$^{1}$, Boyuan Zheng$^{1}$, Junli Wang$^{1}$, Zhiqi Liang$^{1}$, Yuheng Zha$^{1}$, Qiyue Gao$^{1}$, Jixuan Chen$^{1}$, Kun Zhou$^{1}$

\noindent\textbf{Conceptualization and Advising}

Shibo Hao$^{1,\dagger}$, Ziqiao Ma$^{3}$, Zhoujun Cheng$^{1}$, Yu Wang$^{1}$, Tianyang Liu$^{1}$, Feng Yao$^{1}$, Xiaohan Fu$^{7}$, Jingbo Shang$^{1}$, Lianhui Qin$^{1}$, Julian McAuley$^{1}$, Eric P. Xing$^{2}$, Zhengzhong Liu$^{2}$, Rupesh Kumar Srivastava$^{2}$, Zhiting Hu$^{1}$

\medskip
\noindent{\footnotesize Affiliations: $^{1}$ UC San Diego; $^{2}$ MBZUAI IFM; $^{3}$ University of Michigan; $^{4}$ UC Berkeley; $^{5}$ ETH Zurich; $^{6}$ University of Cambridge; $^{7}$ Gray Swan AI.}

\noindent{\footnotesize $^{\dagger}$Corresponding author: \texttt{s5hao@ucsd.edu}.}
\section{\scaffold}

\scaffold Tool Interface. \scaffold exposes a structured tool interface organized into five categories, totaling 39 tools. The browser tools (17 tools) support fine-grained GUI interaction via both low-level pointer events
  (\textsc{browser\_click}, \textsc{browser\_drag\_to}, \textsc{browser\_scroll}) and keyboard input (\textsc{browser\_type}, \textsc{browser\_hotkey}), as well as viewport introspection (\textsc{browser\_screenshot}, \textsc{browser\_get\_viewport\_info}) for screenshot-based visual grounding. The DOM tools (11
   tools) complement this with programmatic access to page structure, including \textsc{dom\_get\_text}, \textsc{dom\_query\_selector}, \textsc{dom\_extract\_links}, and \textsc{dom\_mark\_elements}, enabling agents to read and interact with page content without relying solely on
  pixel-level perception. The file tools (9 tools) cover the full file manipulation workflow—reading, writing, listing, searching, and image reading—while the shell (\textsc{shell\_execute}) and code execution (\textsc{code\_execute}) tools provide terminal
  access and sandboxed Python/JavaScript interpretation. This decomposition ensures that each of the three core capabilities tested by \ours—vision, search, and coding—is supported by dedicated, composable primitives, while keeping the
  interface strongly typed to minimize tool hallucination.

  \begin{table}[t]
  \centering
  \small
  \caption{Complete tool inventory of \textsc{Cocoa-Agent}. Tools are grouped by the three key capabilities defined in \ours: \textit{Vision} (GUI-level browser interaction), \textit{Search} (DOM-level content access and navigation), and \textit{Coding} (code execution, shell commands, and file operations). The \textit{Control} category contains only the task completion signal.}
  \label{tab:cocoa-agent-tools}
  \begin{tabular}{llp{7cm}}
  \toprule
  \textbf{Ability} & \textbf{Tool} & \textbf{Description} \\
  \midrule
  \multirow{13}{*}{Vision}
    & \texttt{browser\_click}           & Click at screen coordinates (left/right/middle; single/double/triple) \\
    & \texttt{browser\_type}            & Type text into the focused element \\
    & \texttt{browser\_press}           & Press a single keyboard key \\
    & \texttt{browser\_hotkey}          & Press a key combination (e.g., \texttt{Ctrl+C}) \\
    & \texttt{browser\_scroll}          & Scroll the page by pixel offset \\
    & \texttt{browser\_move\_to}        & Move the cursor to absolute coordinates \\
    & \texttt{browser\_drag\_to}        & Drag from current position to absolute coordinates \\
    & \texttt{browser\_wait}            & Wait for a specified duration \\
    & \texttt{browser\_screenshot}      & Capture a screenshot of the current viewport \\
    & \texttt{browser\_get\_viewport\_info} & Return current URL and viewport dimensions \\
    & \texttt{image\_read}              & Read an image file and return it as base64 for visual analysis \\
  \midrule
  \multirow{10}{*}{Search}
    & \texttt{browser\_navigate}        & Navigate to a URL (DOM load) \\
    & \texttt{dom\_get\_text}           & Retrieve \texttt{innerText} of the page body \\
    & \texttt{dom\_get\_html}           & Retrieve full page HTML (truncated if long) \\
    & \texttt{dom\_query\_selector}     & Query elements by CSS selector; return tag, id, class, role, etc. \\
    & \texttt{dom\_extract\_links}      & Extract all hyperlinks (text + href), with optional filtering \\
    & \texttt{dom\_mark\_elements}      & Annotate interactive elements with unique BIDs; return element list \\
    & \texttt{dom\_click}               & Click an element by BID \\
    & \texttt{dom\_type}                & Type into an input element by BID \\
    & \texttt{dom\_scroll}              & Scroll an element or the page by BID \\
  \midrule
  \multirow{8}{*}{Coding}
    & \texttt{code\_execute}            & Execute Python or JavaScript in the sandbox; returns stdout/stderr \\
    & \texttt{shell\_execute}           & Execute a shell command and return output \\
    & \texttt{file\_read}               & Read file contents \\
    & \texttt{file\_write}              & Write content to a file \\
  \midrule
  Control
    & \texttt{task\_complete}           & Mark the task as finished and return an optional result string \\
  \bottomrule
  \end{tabular}
\end{table}
\section{Failure Mode Taxonomy}
\label{app:error-taxonomy}
We systematically categorize failures on \ours into three hierarchical layers
based on extensive trajectory analysis across different agentic systems on 153 tasks.
The taxonomy is organized around the \emph{locus of failure}:
whether the root cause resides in the agent's adaptive planning process (E1),
its execution loop (E2), or its vision perceptual grounding in the environment (E3).

\subsection{Type 1 \quad Reasoning \& Planning}
\label{app:error-1}
Planning error arises when the agent's high-level reasoning or decision-making process is fundamentally misaligned with the task objective. These failures occur prior to or independent of execution.

\paragraph{\textbf{E1.1}~Incorrect Reasoning}
The agent fails to construct a valid logical path to the target objective. This takes two forms: 
(1) \emph{Goal displacement}: The agent solves a simplified sub-problem instead of the actual task. 
(2) \emph{Incorrect strategy}: The agent understands the goal but pursues a fundamentally flawed approach, often abandoning core constraints or selecting suboptimal algorithms during execution.

\begin{caseF}{E1.1}{Incorrect Reasoning --- goal displacement}
eight-puzzle-game · \scaffold \par\smallskip
\textbf{Task:} Interact with a live 8-puzzle to discover the hidden goal state, then solve the puzzle from the initial configuration using the minimum number of moves. Crucially, the validation code is path-dependent: reaching the identical final state via a different sequence of moves yields a completely different code.\par\smallskip
\textbf{Trajectory:}\quad The agent successfully probes the API to infer the hidden goal state and computes the optimal 28-move solution via BFS. However, it executes these moves within its ongoing browser session---which already contains prior exploratory steps---and submits the resulting validation code.\\
\textbf{Result:}\quad $\times$\enspace got ``2QGGPD'' / expected ``6CNTDL''\par\smallskip
\textbf{Explanation:}\quad The agent solved the sub-problem (finding the path) but failed the actual task constraint. It did not realize that the environment's validation code strictly requires a clean, optimal-path-only execution, treating "reaching the goal state" as erroneously equivalent to "executing the optimal sequence."
\end{caseF}

\begin{caseF}{E1.1}{Incorrect Reasoning --- missing satisfiability check}
meeting-schedule-constraints · CodeX\par\smallskip
\textbf{Task:} Assign time slots and rooms to four sessions based on constraint files. Crucially, these files detail varying individual participant preferences, containing a mix of overlapping availabilities and strict personal conflicts. Output a valid schedule satisfying everyone, or report that none exists.\par\smallskip
\textbf{Trajectory:}\quad The agent parses the constraints and enumerates assignments using \texttt{itertools.product}. It manages to align several schedules but overlooks the strict conflicts between specific individuals. It halts prematurely and submits a schedule that leaves at least one participant's requirement violated.\\
\textbf{Result:}\quad $\times$\enspace wrong schedule / expected ``(none)''\par\smallskip
\textbf{Explanation:}\quad The individual participant constraints inherently conflict, making a universal schedule impossible (jointly unsatisfiable). The agent employed an incomplete reasoning strategy: it optimized for local agreement but failed to rigorously cross-check the personal conflicts among all individuals, missing the fact that the only correct answer is an empty schedule.
\end{caseF}

\paragraph{\textbf{E1.2}~Imprecision}
The agent executes the correct high-level procedure but returns an incorrect value due to execution-level inaccuracies. This manifests in two ways:
In the \emph{precision} variant, floating-point accumulation or premature rounding introduces a small but evaluation-critical offset to an otherwise sound computation.
In the \emph{scope} variant, the agent applies the correct algorithm to the wrong data boundary, such as failing to filter out items that should be excluded (e.g., counting appendix citations alongside the main body).

\begin{caseF}{E1.2}{Imprecision --- floating-point accumulation}
yearly-costco-gas-receipts · CodeX\par\smallskip
\textbf{Task:} Compute the weighted average gas price across a full year of Costco gas receipts, rounding the final output to exactly three decimal places.\par\smallskip
\textbf{Trajectory:}\quad The agent correctly identifies all 2023 receipts and formulates the right price-volume product logic. However, it iteratively sums these values using intermediate rounding rather than maintaining full floating-point precision until the final division.\\
\textbf{Result:}\quad $\times$\enspace got ``4.217'' / expected ``4.216''\par\smallskip
\textbf{Explanation:}\quad The high-level mathematical procedure is completely sound, but the execution suffers from imprecision. By failing to defer rounding to the absolute final step, floating-point accumulation introduces a tiny but evaluation-critical offset of 0.001.
\end{caseF}

\begin{caseF}{E1.2}{Imprecision --- scope boundary}
phd-student-paper-analysis · \scaffold \par\smallskip
\textbf{Task:} Count how many times specific top-cited references appear strictly within the \textbf{main body} of a PhD thesis, explicitly excluding any appendices.\par\smallskip
\textbf{Trajectory:}\quad The agent parses the PDF and correctly identifies the target citation string (e.g., "Saparov and He, 2022"). It then executes a global text search across the entire document, tallying all matches without applying any structural or section-based filtering.\\
\textbf{Result:}\quad $\times$\enspace got 5 / expected 4\par\smallskip
\textbf{Explanation:}\quad The agent applied the correct counting algorithm but to the wrong data boundary. By blindly scanning the full document, it erroneously included citations located in the appendix, silently corrupting the final count due to a scope mis-specification.
\end{caseF}

\paragraph{\textbf{E1.3}~Format Error}
The agent derives the correct answer but fails to structure it for the evaluator. This occurs via:
(1) \emph{Tag omission}: Burying the valid answer in prose without the required \texttt{<answer>} tags.
(2) \emph{Partial delivery}: Submitting only a subset of a required multi-part output, treating the first completed field as the entire answer.

\begin{caseF}{E1.3}{Format Error --- partial delivery}
tableau-profit-margin-analysis · CodeX\par\smallskip
\textbf{Task:} Given a Tableau dashboard, extract two insights for a specified region: (a) the month with the highest profit margin, and (b) the year with the most volatile profit margins.\par\smallskip
\textbf{Trajectory:}\quad The agent successfully scrapes the underlying dashboard data, computing both the monthly profit margins and the yearly standard deviations. However, immediately upon resolving the first sub-question, it prematurely terminates the run and submits only the highest-margin month (May, 26.0\%), silently discarding its volatility computations.\\
\textbf{Result:}\quad $\times$\enspace got correct answer to (a) only / expected answers to (a) and (b)\par\smallskip
\textbf{Explanation:}\quad The agent's analytical reasoning and data extraction were flawless, but it failed on output formatting constraints. By treating the first resolved sub-query as the completion of the entire prompt, it committed a partial delivery error, leaving the second requirement unfulfilled despite already possessing the correct data.
\end{caseF}

\subsection{Type 2 \quad Tool \& Execution}
\label{app:error-2}

While the high-level plan may be sound, Type 2 failures emerge as structural breakdowns in the agent's active interaction loop. The task halts not due to bad logic, but through poor execution: tool misuse, absent recovery mechanisms, or behavioral stagnation.

\paragraph{\textbf{E2.1}~Infinite Loop}
The agent fails to self-correct and falls into an endless execution cycle. This manifests in two primary ways:
(1) \emph{Repetition}: Blindly re-issuing identical actions despite receiving consistent error signals.
(2) \emph{Exhaustion}: Endlessly tweaking low-level parameters (e.g., scrolling, resizing) without realizing the overarching strategy is doomed.
Crucially, this includes environment-induced loops (e.g., broken tool APIs), as the underlying failure is the agent's inability to recognize and escalate a systemic block.

\begin{caseF}{E2.1}{Infinite Loop --- exhaustion variant}
nonogram · Claude Code\par\smallskip
\textbf{Task:} Solve a nonogram puzzle provided as an image and identify the specific letter formed by the completed grid.\par\smallskip
\textbf{Trajectory:}\quad The agent attempts to transcribe the puzzle's numerical clues using a low-level visual extraction strategy. It expends its entire 51-turn budget iteratively cropping and reading tiny row and column regions via PIL, endlessly tweaking its focus area but never progressing to the actual solving phase.\\
\textbf{Result:}\quad $\times$\enspace got (none) / expected ``w''\par\smallskip
\textbf{Explanation:}\quad The agent became trapped in an exhaustion loop. Instead of recognizing the severe inefficiency of manual pixel-level transcription and pivoting to a higher-level abstraction (e.g., passing the full image to a robust OCR tool once and writing a programmatic constraint solver).
\end{caseF}

\paragraph{\textbf{E2.2}~Anti-Bot Barriers}
The agent is blocked by websites' security mechanisms but fails to recognize the interruption. It misinterprets the interstitial bot-detection page as the actual target content, resulting in hallucinated answers or silent failures without user escalation.

\begin{caseF}{E2.2}{Anti-Bot Barriers}
neurips-citation-analysis · CodeX\par\smallskip
\textbf{Task:} Retrieve current citation counts for specific NeurIPS papers using public academic indexing services.\par\smallskip
\textbf{Trajectory:}\quad The agent attempts to scrape an academic database but is intercepted by a Cloudflare bot-detection challenge. Failing to recognize the HTML response as a security barrier, it blindly parses the challenge page as if it were the target paper's record. Finding no citation metrics within the boilerplate security warning, it confidently outputs a citation count of zero.\\
\textbf{Result:}\quad $\times$\enspace got 0 / expected actual citation count\par\smallskip
\textbf{Explanation:}\quad The agent's failure stems entirely from its inability to contextualize interstitial network obstacles. Instead of detecting the anti-bot barrier and pivoting its retrieval strategy (or safely aborting), it treated the security interception as a valid, empty search result. This lack of situational awareness directly transformed a simple tool-access block into a confidently hallucinated factual error.
\end{caseF}

\paragraph{\textbf{E2.3}~Tool Result Hallucination}
The agent proceeds based on fabricated outputs or a corrupted memory state. 
(1) \emph{Tool hallucination}: Invoking non-existent tools or fabricating execution results without actually issuing a call. 
(2) \emph{Context truncation}: As the interaction history exceeds the context window, early instructions or critical findings are silently dropped. The agent continues unawares, leading to repetitive actions and progressive decoupling from the core objective.

\begin{caseF}{E2.3}{Tool Result Hallucination --- context truncation}
glp1-pharmaceutical-analysis · \scaffold\par\smallskip
\textbf{Task:} Extract and systematically compare primary endpoint data for GLP-1 drug trials across a dense corpus of 47 research papers.\par\smallskip
\textbf{Trajectory:}\quad The agent initiates a systematic review, successfully retrieving and extracting data from several papers in the early turns. However, as the dense action-observation history exceeds the context window limits, these early findings are silently evicted (compacted). Unaware of this memory loss, the agent repeatedly notes that prior results are "missing," prompting it to re-fetch the exact same endpoint data and restart sub-tasks from scratch. This endless cycle persists until the run halts at turn 71.\par\smallskip
\textbf{Result:}\quad $\times$\enspace got (incomplete) / expected full comparative analysis\par\smallskip
\textbf{Explanation:}\quad This is a textbook manifestation of context truncation. The agent's logic did not fail; rather, it was forced to operate on a degraded, amnesic memory state. Because critical early findings were dropped to accommodate new observations, the agent lost its global state awareness. Instead of synthesizing building blocks into a final comparison, it became trapped in a Sisyphean cycle of redundant retrieval, progressively decoupling from the overarching objective.
\end{caseF}

\subsection{Type 3 \quad Visual Grounding}
\label{app:error-3}
While Type 1 and Type 2 failures stem from flawed strategy or broken interaction loops, Type 3 errors represent a fundamental perceptual disconnect. Here, the agent may reason correctly and execute actions smoothly, but ultimately fails because it misreads, overlooks, or cannot semantically map the visual state of the environment.

\paragraph{\textbf{E3.1}~Visual Detail}
The agent captures the global layout of a scene but fails to accurately resolve fine-grained visual features. This deficiency primarily manifests in three areas: \emph{small-target detection} (missing or mislocalizing tiny icons and dense UI elements), \emph{text recognition} (misreading small, low-contrast, or stylized fonts), and \emph{thin-object perception} (misjudging the alignment, boundaries, or intersections of lines, arrows, and borders).

\begin{caseF}{E3.1}{Visual Detail --- measurement imprecision}
mountain-video · CodeX\par\smallskip
\textbf{Task:} Watch a mountain-climbing video and extract the exact elevation of each labeled step shown on the route diagram.\par\smallskip
\textbf{Trajectory:}\quad The agent successfully isolates the relevant video frame and applies both its Vision-Language Model (VLM) capabilities and OCR tools to parse the on-screen text. While it correctly reads the surrounding elevation fields, it misreads a single digit in the third-step marker's label, extracting 8710\,m instead of the true value.\\
\textbf{Result:}\quad $\times$\enspace got 8710\,m / expected 8690\,m\par\smallskip
\textbf{Explanation:}\quad This failure highlights a critical deficit in fine-grained visual acuity. Even when directly combining holistic VLM perception with dedicated OCR processing, the agent failed to accurately resolve low-level pixel details. Misinterpreting just one digit corrupted the extracted measurement, shifting the final answer by 20\,m and pushing it beyond the acceptable evaluation tolerance.
\end{caseF}

\paragraph{\textbf{E3.2}~Visual Knowledge}
The agent forms a correct visual representation of the scene but lacks the prior world knowledge to map these features to their corresponding textual concepts. The breakdown is semantic rather than perceptual: the agent clearly "sees" the depicted entities (e.g., specific people, locations, turn signal, or cultural artifacts) but cannot identify them due to insufficient parametric knowledge.

\begin{caseF}{E3.2}{Visual Knowledge --- color naming}
taylor-swift-eras-tour · Claude Code\par\smallskip
\textbf{Task:} Identify the dominant color of a set of Eras Tour album covers and map it to the closest standard color name.\par\smallskip
\textbf{Trajectory:}\quad The agent successfully samples the image palette and computes a mathematically precise RGB mean of (148, 103, 189). However, when converting this data into a semantic label, it abruptly classifies this clearly purplish coordinate as "gray" and submits the wrong answer.\\
\textbf{Result:}\quad $\times$\enspace got ``gray'' / expected ``purple''\par\smallskip
\textbf{Explanation:}\quad The failure represents a fundamental gap in parametric world knowledge rather than perception. The agent flawlessly extracted the visual representation (the RGB values) but failed to align this data with the correct textual concept, demonstrating a breakdown in visual-to-semantic mapping.
\end{caseF}

\begin{caseF}{E3.2}{Visual Knowledge --- person misidentification}
grammy-awards-singer-photo · CodeX\par\smallskip
\textbf{Task:} Identify the two specific artists pictured together in a Grammy ceremony photo, then calculate the total combined number of awards they won between 2020 and 2023 (inclusive).\par\smallskip
\textbf{Trajectory:}\quad The agent correctly identifies Billie Eilish on the right but misidentifies Lana Del Rey on the left as "Taylor Swift." Proceeding with this flawed visual extraction, it queries the 2020--2023 award histories for Eilish and Swift, summing them to arrive at an incorrect final count.\\
\textbf{Result:}\quad $\times$\enspace got 3 / expected 7\par\smallskip
\textbf{Explanation:}\quad The numerical calculation error is merely a downstream symptom of an epistemic visual breakdown. The agent successfully perceived the facial features but lacked the parametric knowledge to map them to Lana Del Rey, defaulting to a superficially similar high-frequency concept (Taylor Swift). This initial visual-to-semantic misidentification poisoned the subsequent data retrieval, dooming the entire multi-hop task.
\end{caseF}

\begin{caseF}{E3.2}{Visual Knowledge --- directional convention}
bear-statue-building-direction · CodeX\par\smallskip
\textbf{Task:} Determine the cardinal direction a building faces, given a street-view image of a bear statue at a roundabout and the navigation instruction: ``enter the roundabout and exit at the first exit.''\par\smallskip
\textbf{Trajectory:}\quad The agent correctly observes the physical layout of the roundabout and the orientation of the statue. To deduce the path, however, it blindly applies a left-hand-traffic (right-hand drive) heuristic. Assuming traffic flows clockwise, it takes the immediate left branch as the "first exit," failing to recognize from the visual context that the scene is in a right-hand-traffic country.\\
\textbf{Result:}\quad $\times$\enspace got ``northwest'' / expected ``northeast''\par\smallskip
\textbf{Explanation:}\quad This is a failure of contextual world knowledge. The agent correctly "read" the visual geometry but applied the wrong geographic prior. By erroneously defaulting to a left-hand-traffic convention, it simulated the roundabout navigation backwards (clockwise instead of counterclockwise), reversing the vehicle's bearing and ultimately producing the wrong cardinal direction.
\end{caseF}

\paragraph{\textbf{E3.3}~Missing Visual Perception}
The agent reads the page via DOM queries rather than inspecting the rendered
visual output, causing it to miss content that exists only in the pixel buffer.
Modern web applications frequently render data through \texttt{<canvas>}
elements, SVG overlays, or JavaScript-driven frameworks that expose no
semantic content to the DOM; the agent either reports such content as absent
or fabricates plausible values from contextual priors.

\begin{caseF}{E3.3}{Missing Visual Perception --- DOM opacity}
tableau-profit-margin-analysis · CodeX\par\smallskip
\textbf{Task:} Read a live Tableau dashboard and visually extract the month with the highest profit margin and the year with the most volatile margins for a given region.\par\smallskip
\textbf{Trajectory:}\quad Attempting to bypass visual processing entirely, the agent invokes a backend scraper API (\texttt{tableauscraper}) to pull the raw underlying data fields. Relying strictly on this textual payload, it identifies "May" at "26.0\%" and confidently submits it without ever inspecting the rendered dashboard pixels.\\
\textbf{Result:}\quad $\times$\enspace got May / 26.0\% / expected March / 15.3\%\par\smallskip
\textbf{Explanation:}\quad The agent fell victim to a semantic "mirage." Modern BI tools like Tableau perform complex client-side aggregations that only manifest in the final visual render. By relying exclusively on the raw, unaggregated API payload rather than the visual ground truth, the agent extracted mathematically correct but visually absent data, completely failing the inherently perceptual task.
\end{caseF}

\begin{caseF}{E3.3}{Missing Visual Perception --- SVG measurement mismatch}
quadrant-visual-issues · CodeX\par\smallskip
\textbf{Task:} Render a Mermaid quadrant chart, identify visual issues such as overlapping labels (Q2), and find the minimum \texttt{chartWidth} (in multiples of 100) that visually resolves all overlaps (Q3).\par\smallskip
\textbf{Trajectory:}\quad The agent renders the chart using Playwright but attempts to detect label collisions by querying DOM bounding boxes (\texttt{getBoundingClientRect()}) rather than employing actual visual inspection. The DOM queries falsely report that overlaps persist at a width of 800\,px. Trusting this API over its own visual capabilities, the agent needlessly widens the chart until the DOM clears the overlaps at 1200\,px.\\
\textbf{Result:}\quad $\times$\enspace got Q3\,=\,1200 / expected Q3\,=\,800 (Q1 and Q2 were correct)\par\smallskip
\textbf{Explanation:}\quad The agent attempted to solve a strictly visual geometry problem using non-visual DOM structures. In complex SVGs, structural bounding boxes often overlap even when the rendered text pixels do not. By blindly trusting this programmatic mirage over the actual pixel-level rendering, which clearly showed overlaps were resolved at 800\,px—the agent overcompensated and submitted a grossly inflated width.
\end{caseF}


\subsection{More Error Analysis}
\label{app:more-error-analysis}

\begin{figure}[h]
    \centering
    \vspace{-1em}
    \includegraphics[width=0.80\linewidth]{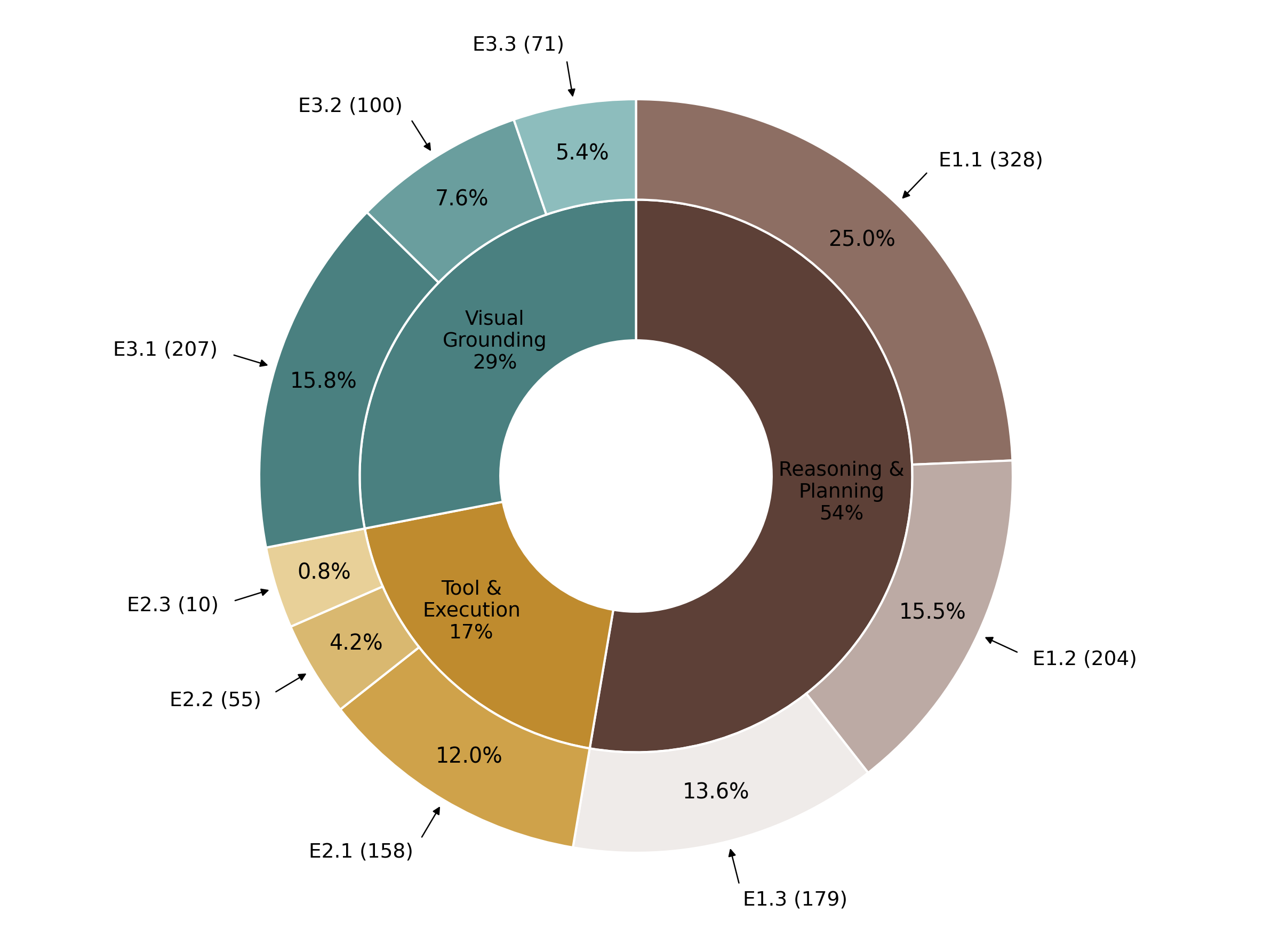}
    \vspace{-0.5em}
    \caption{
        Aggregate failure distribution across all 6 models evaluated under
        \scaffold on \ours.
        The inner ring: Reasoning \& Planning (54\%), Tool \& Execution
        (17\%), Visual Grounding (29\%).
        The outer ring details all 9 active subcategories (E1.1--E3.3).
        Displayed percentages reflect each subcategory's share of total
        failure-mode mentions across all 722 failed runs.
    }
    \label{app:fig:err-cocoa-all}
\end{figure}

\begin{figure}[h]
    \centering
    \vspace{-1em}
    \includegraphics[width=0.80\linewidth]{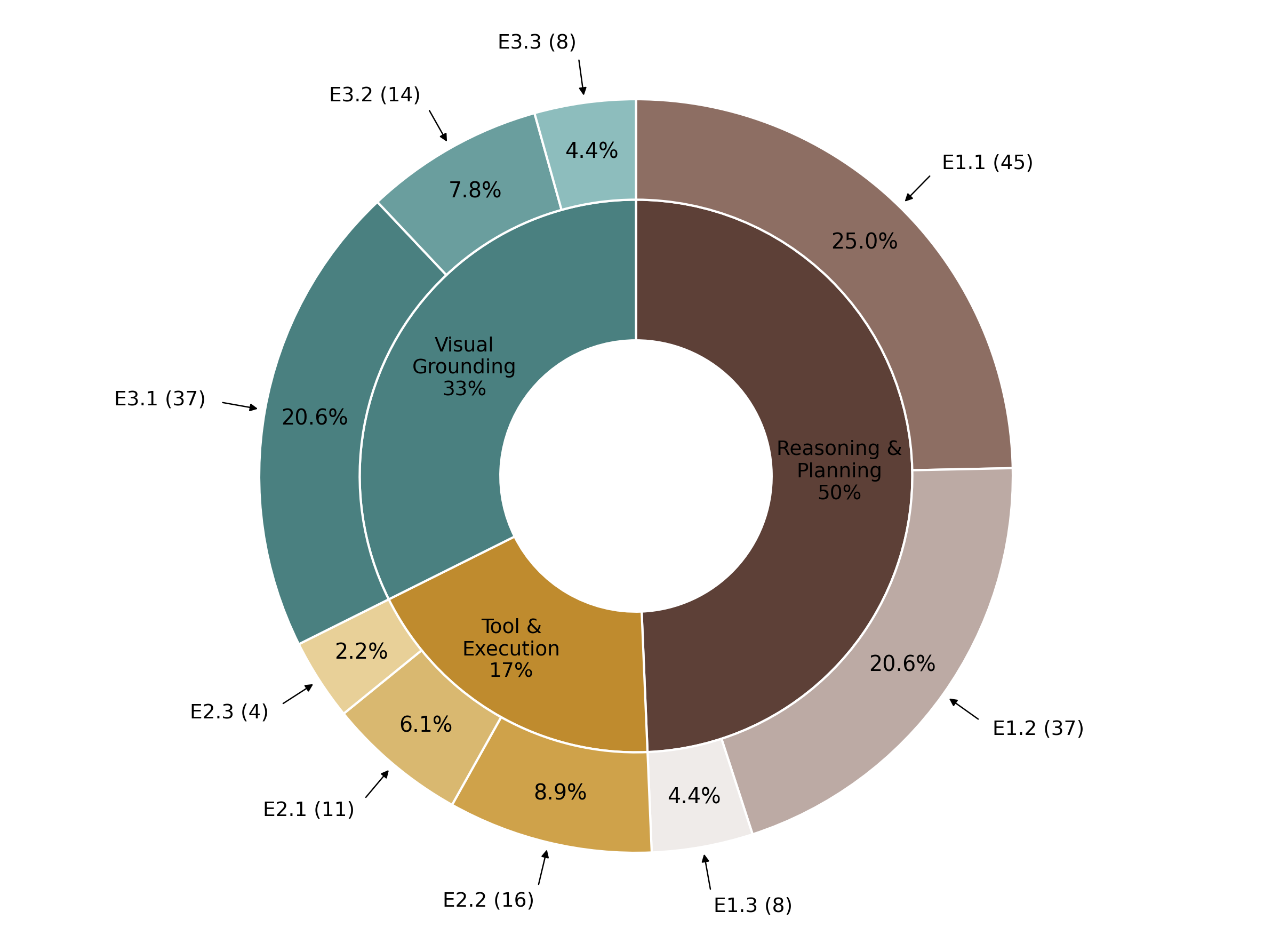}
    \vspace{-0.5em}
    \caption{
        \scaffold with Claude Sonnet 4.6 failures on \ours mapped to the
        failure taxonomy.
        The inner ring: Reasoning \& Planning (50\%), Tool \& Execution
        (17\%), Visual Grounding (33\%).
        The outer ring details 9 active subcategories.
        Displayed percentages reflect each subcategory's share of total
        failure-mode mentions across all 114 failed runs.
    }
    \label{app:fig:err-claude-sonnet}
\end{figure}

\begin{figure}[h]
    \centering
    \vspace{-1em}
    \includegraphics[width=0.80\linewidth]{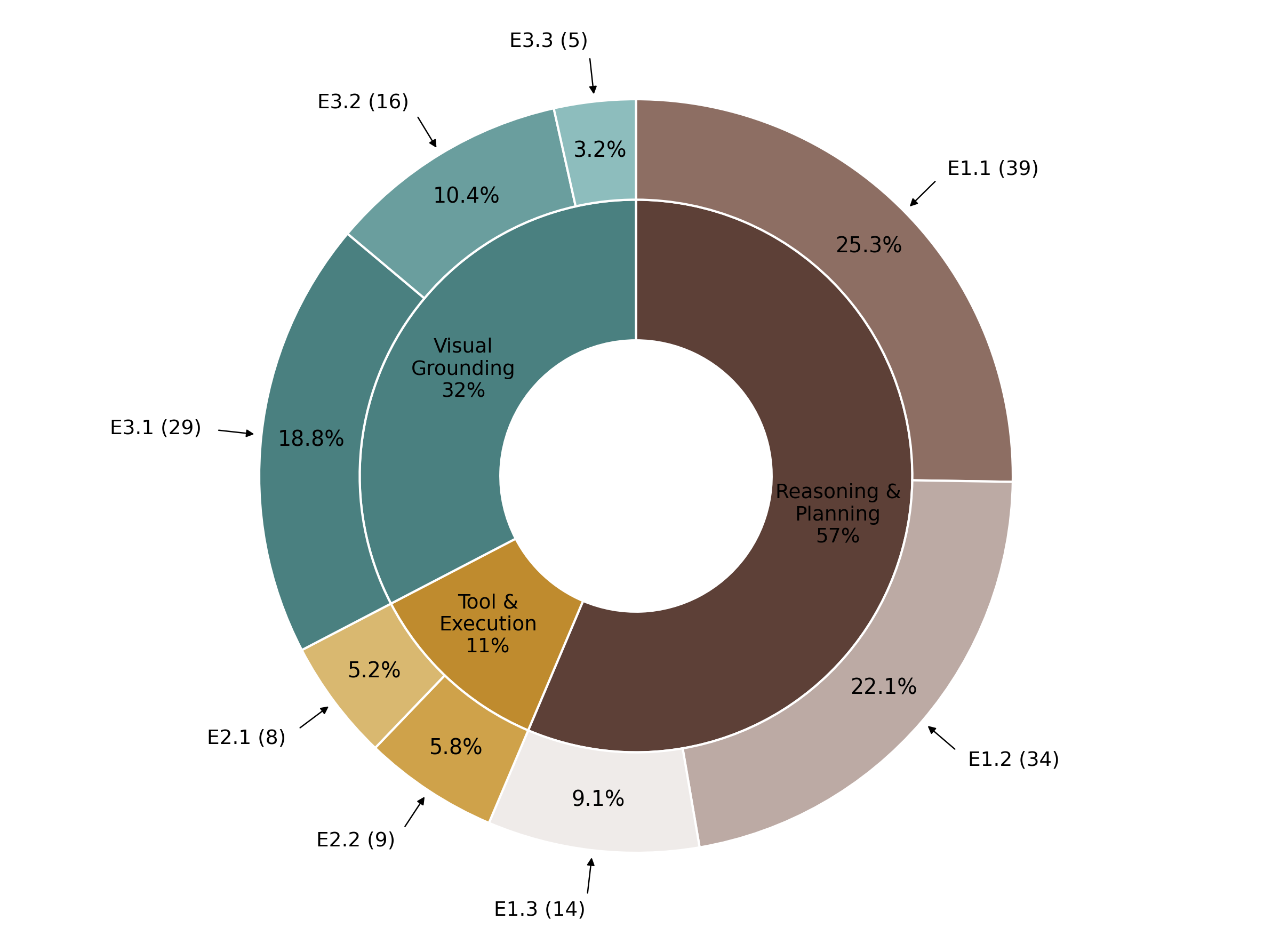}
    \vspace{-0.5em}
    \caption{
        \scaffold with GPT-5.4 failures on \ours mapped to the failure
        taxonomy.
        The inner ring: Reasoning \& Planning (57\%), Tool \& Execution
        (11\%), Visual Grounding (32\%).
        The outer ring details 8 active subcategories.
        Displayed percentages reflect each subcategory's share of the total
        failure-mode mentions across all 96 failed runs.
    }
    \label{app:fig:err-gpt54}
\end{figure}

\begin{figure}[h]
    \centering
    \vspace{-1em}
    \includegraphics[width=0.80\linewidth]{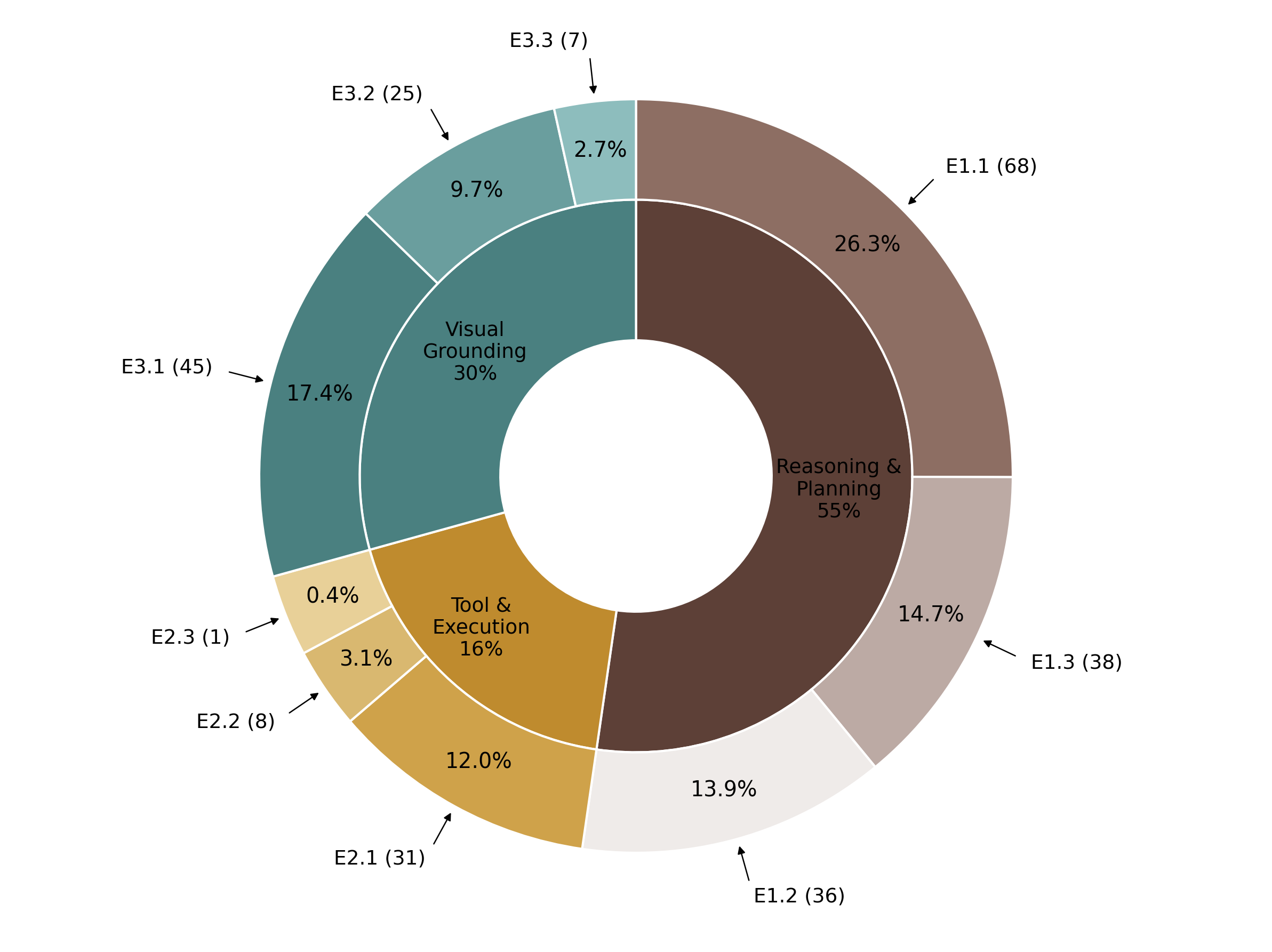}
    \vspace{-0.5em}
    \caption{
        \scaffold with Kimi-k2.5 failures on \ours mapped to the failure
        taxonomy.
        The inner ring: Reasoning \& Planning (55\%), Tool \& Execution
        (15\%), Visual Grounding (30\%).
        The outer ring details 9 active subcategories.
        Displayed percentages reflect each subcategory's share of the total
        failure-mode mentions across all 137 failed runs.
    }
    \label{app:fig:err-kimi}
\end{figure}

\begin{figure}[h]
    \centering
    \vspace{-1em}
    \includegraphics[width=0.80\linewidth]{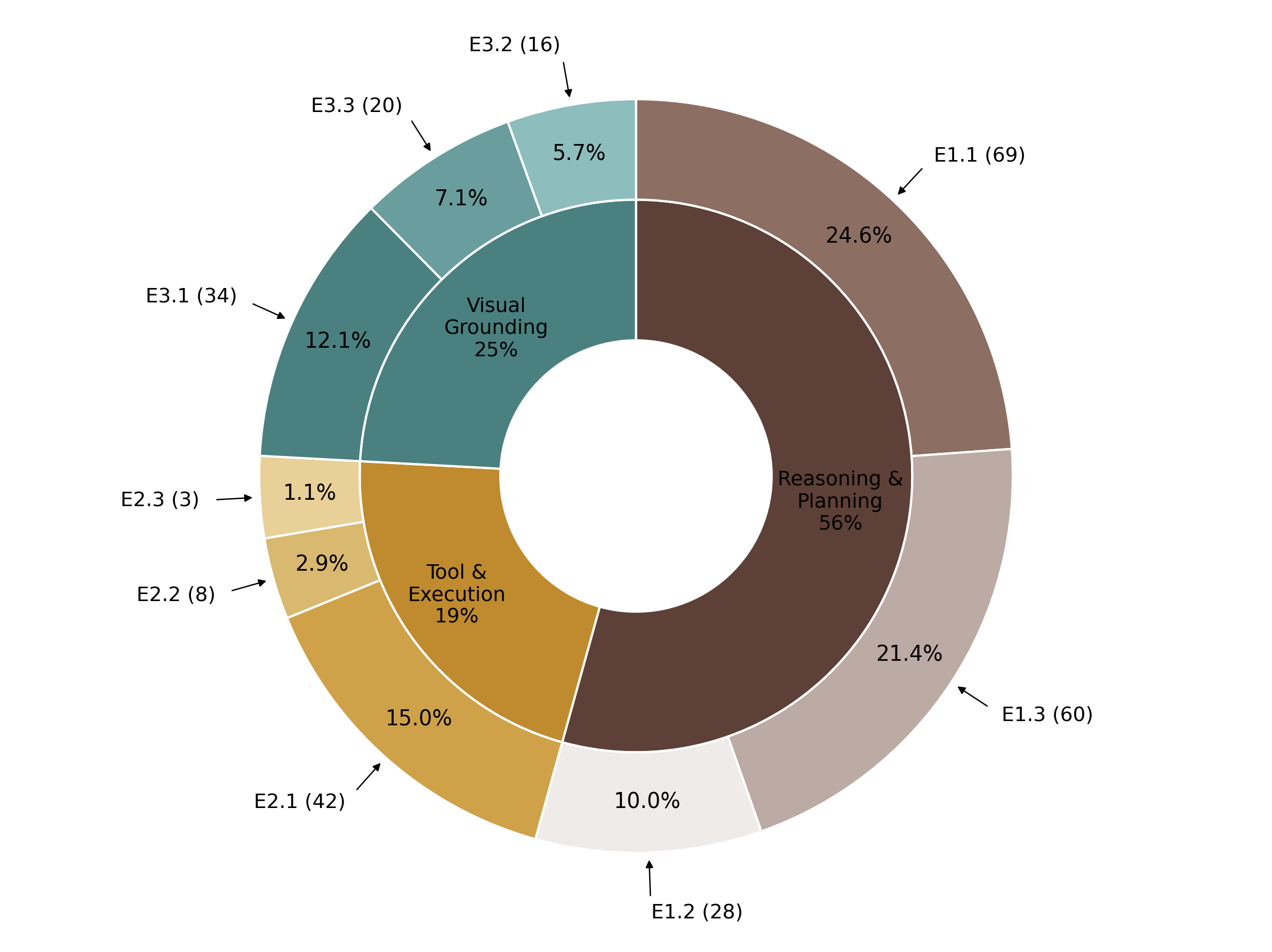}
    \vspace{-0.5em}
    \caption{
        \scaffold with Qwen3.5-397B failures on \ours mapped to the failure
        taxonomy.
        The inner ring: Reasoning \& Planning (56\%), Tool \& Execution
        (19\%), Visual Grounding (25\%).
        The outer ring details 9 active subcategories.
        Displayed percentages reflect each subcategory's share of the total
        failure-mode mentions across all 139 failed runs.
    }
    \label{app:fig:err-qwen}
\end{figure}

\begin{figure}[h]
    \centering
    \vspace{-1em}
    \includegraphics[width=0.80\linewidth]{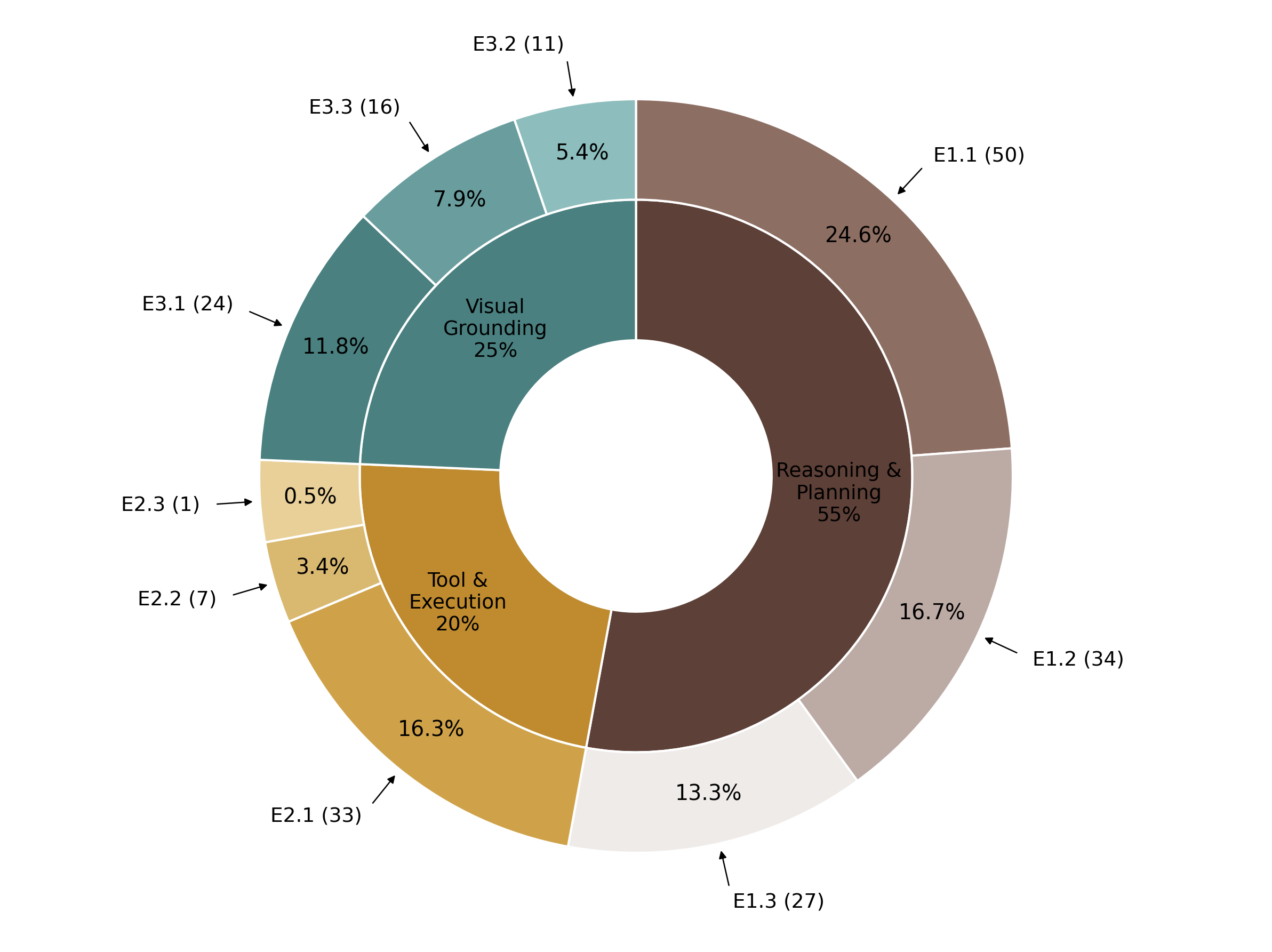}
    \vspace{-0.5em}
    \caption{
        \scaffold with Gemini 3.1 Pro Thinking failures on \ours mapped to the
        failure taxonomy.
        The inner ring: Reasoning \& Planning (55\%), Tool \& Execution
        (20\%), Visual Grounding (25\%).
        The outer ring details 9 active subcategories.
        Displayed percentages reflect each subcategory's share of the total failure-mode mentions across all 107 failed runs.
    }
    \label{app:fig:err-gemini-thinking}
\end{figure}

\begin{figure}[h]
    \centering
    \vspace{-1em}
    \includegraphics[width=0.80\linewidth]{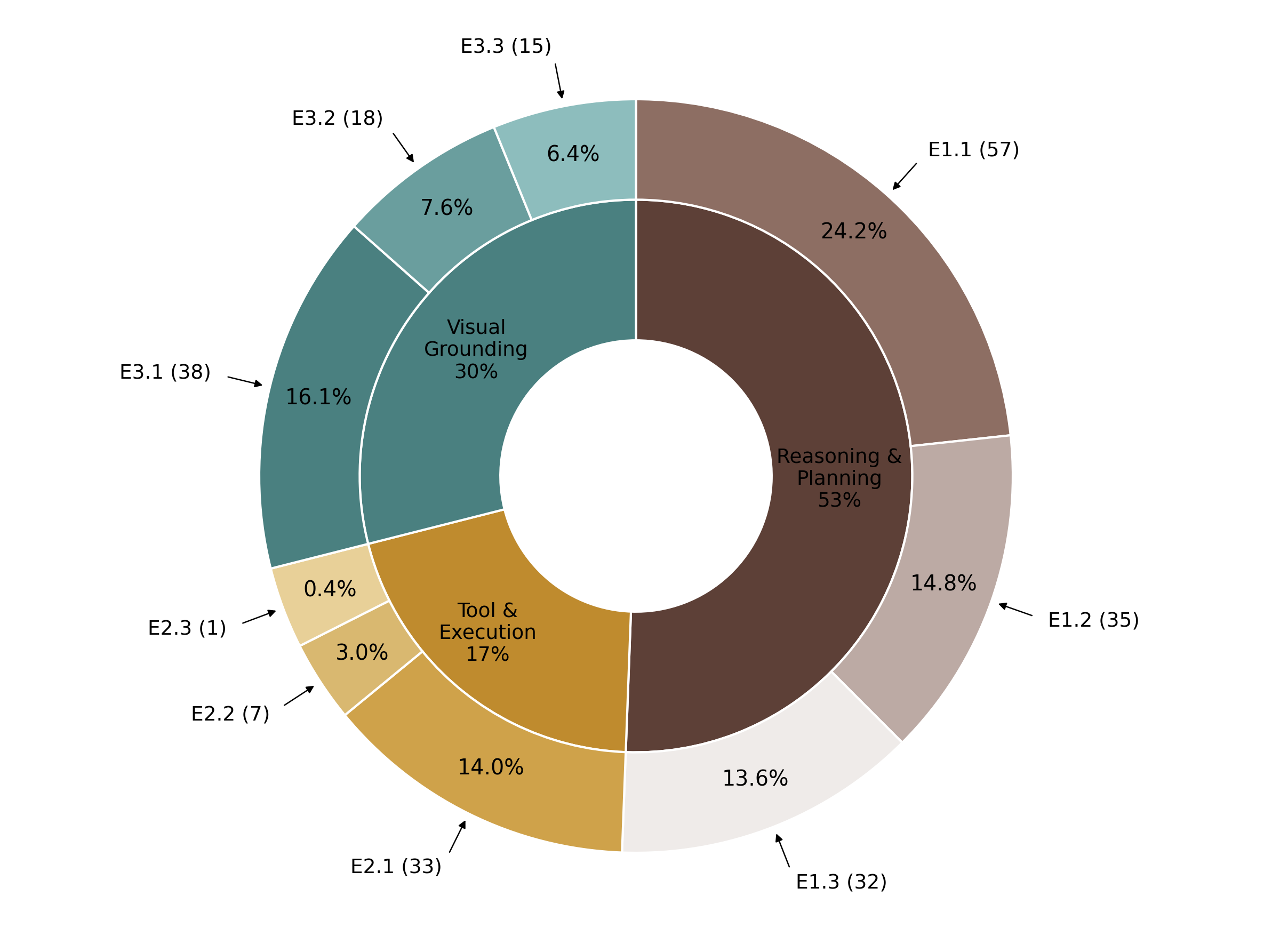}
    \vspace{-0.5em}
    \caption{
        \scaffold with Gemini 3 Flash failures on \ours mapped to the failure
        taxonomy.
        The inner ring: Reasoning \& Planning (53\%), Tool \& Execution
        (17\%), Visual Grounding (30\%).
        The outer ring details 9 active subcategories.
        Displayed percentages reflect each subcategory's share of the total
        failure-mode mentions across all 125 failed runs.
    }
    \label{app:fig:err-gemini-flash}
\end{figure}

\begin{figure}[h]
    \centering
    \vspace{-1em}
    \includegraphics[width=0.80\linewidth]{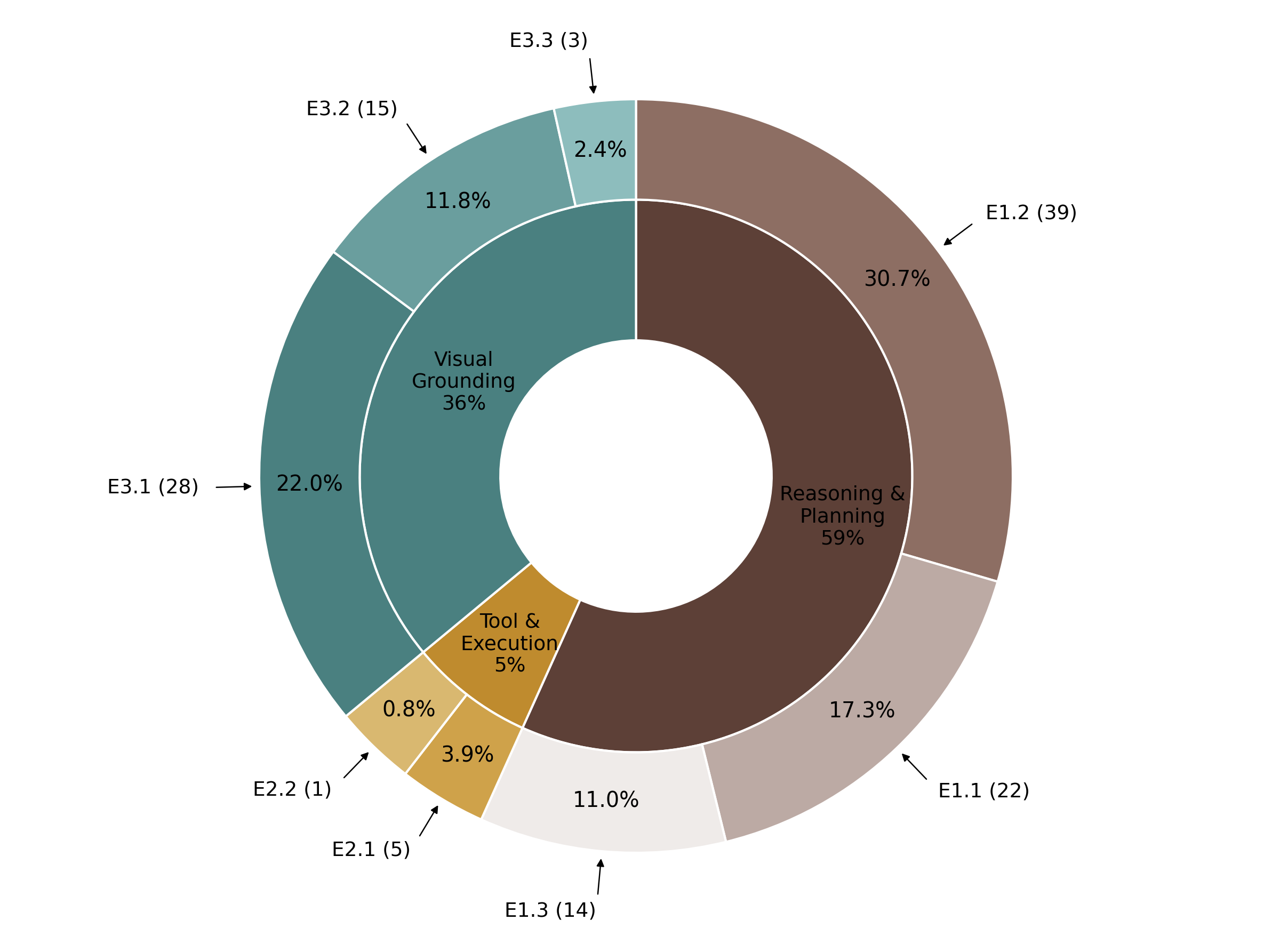}
    \vspace{-0.5em}
    \caption{
        OpenAI Codex failures on \ours mapped to the failure taxonomy.
        The inner ring: Reasoning \& Planning (59\%), Tool \& Execution
        (5\%), Visual Grounding (36\%).
        The outer ring details 8 active subcategories.
        Displayed percentages reflect each subcategory's share of total
        failure-mode mentions across all 86 failed runs.
    }
    \label{app:fig:err-codex}
\end{figure}

Figures~\ref{app:fig:err-cocoa-all}--\ref{app:fig:err-codex} present the full
failure-mode breakdown for every model and scaffold configuration evaluated
on \ours.
Each donut chart uses the revised three-tier taxonomy described in
Section~\ref{app:error-1}--~\ref{app:error-3}: the inner ring shows the aggregate share
of \textbf{Reasoning \& Planning} (E1, brown),
\textbf{Tool \& Execution} (E2, gold), and
\textbf{Visual Grounding} (E3, teal);
the outer ring details the active leaf subcategories (E1.1--E3.3), with arcs
shaded from dark to light in descending frequency within each group, and
parenthesized values indicating raw occurrence counts.
Displayed percentages in the outer ring reflect each subcategory's share of
\emph{total failure-mode mentions} (i.e.\ the sum of all Ex.x codes assigned
across all failed runs); a single trajectory may contribute multiple failure modes.

\clearpage

\subsection{LLM-as-Judge: Error Classification Prompt}
\label{app:judge-prompt}

We use an LLM-as-judge ({claude-sonnet-4-6})
to assign one or more failure-mode codes from the Error taxonomy
(Sections~\ref{app:error-1}--\ref{app:error-3}) to each failed trajectory.
The judge receives five inputs per run: the task description (README), the
evaluator's expected answer, a run summary (status, iteration count,
agent answer, evaluator feedback), and a compact text trace reconstructed
from the execution log---screenshots and base-64 blobs are removed, and per-action observations are truncated to a fixed character budget by action
type (e.g.\ 400\,chars for DOM element lists, 0\,chars for pure screenshots,
600\,chars default).
The full prompt is reproduced below; the taxonomy section mirrors the
definitions in Sections~\ref{app:error-1}--\ref{app:error-3} exactly.

The judge is instructed to \emph{include} a category whenever the failure
\emph{partially or primarily} matches its description, erring on the side of inclusion; a single trajectory may receive multiple codes.
A free-text \texttt{OTHER:<phrase>} escape is provided for failures that
genuinely fall outside the taxonomy, though in practice fewer than 0.5\%
of classified runs use it.
The judge is asked to reply with \emph{only} a comma-separated list of codes
(e.g.\ \texttt{E1.1,~E3.3}) so that the output is unambiguously parsable with
a short regex; no chain-of-thought is elicited and no system prompt is used.

\begin{caseF}{Prompt}{LLM-as-judge classification prompt}
\texttt{[Role]}\par\smallskip
\texttt{You are an expert AI benchmark analyst. A benchmark agent has failed the
task below. Classify this failure into one or more categories from the taxonomy
provided.}\par\medskip
\texttt{[Classification guidelines]}\par\smallskip
\texttt{- Apply a category if the failure partially or primarily matches its
description. Err on the side of inclusion rather than exclusion.}\\
\texttt{- A single run may exhibit multiple failure modes; list all that apply.}\\
\texttt{- If no existing Ex.x captures the core failure, append
OTHER:<one short phrase>. Only use OTHER when truly nothing fits.}\par\medskip
\texttt{[Taxonomy]}\par\smallskip
\texttt{Type 1 --- Reasoning \& Planning}\\
\texttt{E1.1 Incorrect Reasoning / E1.2 Imprecision / E1.3 Format Error}\par\smallskip
\texttt{Type 2 --- Tool \& Execution}\\
\texttt{E2.1 Infinite Loop / E2.2 Anti-Bot Barriers / E2.3 Tool Result Hallucination}\par\smallskip
\texttt{Type 3 --- Visual Grounding}\\
\texttt{E3.1 Visual Detail / E3.2 Visual Knowledge / E3.3 Missing Visual Perception}\par\medskip
\texttt{[Per-run context]}\par\smallskip
\texttt{Task: <task\_name>}\par\smallskip
\texttt{\#\#\# Task Instruction}\\
\texttt{<full README.md of the task>}\par\smallskip
\texttt{\#\#\# Scoring / Expected Answer}\\
\texttt{<EXPECTED = ... lines extracted from test.py>}\par\smallskip
\texttt{\#\#\# Run Summary}\\
\texttt{- Run status   : <passed | failed | error>}\\
\texttt{- Iterations   : <N>}\\
\texttt{- Agent answer : <task\_result or answer field>}\\
\texttt{- Eval feedback: <evaluator feedback string>}\par\smallskip
\texttt{\#\#\# Agent Action Trace (screenshots removed; observations truncated)}\\
\texttt{<compact trace: [N] THINK: ... / ACTION: ... / OBS: ...>}\\
\texttt{<total capped at 1M chars via head+tail truncation>}\par\medskip
\texttt{[Output instruction]}\par\smallskip
\texttt{Which failure categories apply to this run?}\\
\texttt{Reply with ONLY a comma-separated list of Ex.x codes (and optionally}\\
\texttt{OTHER:<short-phrase>). Nothing else.}\\
\texttt{Example: E1.1, E3.3, OTHER:dependency-missing}
\end{caseF}

\paragraph{Observation truncation budget.}
To keep the trace within the model's effective context while preserving the
most diagnostic signal, observations are truncated per action type before the
trace is assembled (Table~\ref{tab:obs-budget}).
Truncation is applied as a \emph{head} slice (the first $L$ characters), so
the opening structure of each observation---page title, status code, element
count---is always retained.
The total trace is then hard-capped at $10^6$ characters using a symmetric
head+tail window; in practice, fewer than 2\% of runs hit this cap.

\begin{table}[htbp]
\centering
\caption{Observation truncation limits per action type used when building the judge prompt.}
\label{tab:obs-budget}
\small
\begin{tabular}{@{}lrl@{}}
\toprule
Action Type & Limit (chars) & Rationale \\
\midrule
\multicolumn{3}{@{}l}{\textit{DOM \& Navigation (Prone to long lists; only page context needed)}} \\
\quad \texttt{dom\_get\_text}               & 500 & \\
\quad \texttt{dom\_mark\_elements}          & 400 & \\
\quad \texttt{dom\_get\_html}               & 300 & \\
\quad \texttt{browser\_navigate}            & 300 & \\
\quad \texttt{browser\_get\_viewport\_info} & 200 & \\
\midrule
\multicolumn{3}{@{}l}{\textit{Visual \& Binary Data (Non-textual data; omitted entirely)}} \\
\quad \texttt{browser\_screenshot}          & 0   & Pure image data \\
\quad \texttt{image\_read}                  & 0   & Base-64 encoded blob \\
\midrule
(All other actions)                         & 600 & Default text limit \\
\bottomrule
\end{tabular}
\end{table}

\end{document}